\documentclass{article}

\usepackage[final,nonatbib]{neurips_2022}

\usepackage[utf8]{inputenc} 
\usepackage[T1]{fontenc}    
\usepackage{hyperref}       
\usepackage{url}            
\usepackage{booktabs}       
\usepackage{amsfonts}       
\usepackage{nicefrac}       
\usepackage{microtype}      
\usepackage{xcolor}         
\usepackage{comment}
\usepackage{algorithm}
\usepackage{algpseudocode}
\usepackage[numbers]{natbib}

\usepackage{hyperref}
\usepackage{url}

\def\ColorFive{rgb:yellow,0;blue,0;white,5;black,10;green,0;red,0;orange,0}

\usepackage{graphicx}
\usepackage[export]{adjustbox}
\usepackage[font=small]{caption}
\usepackage[labelformat = empty,position=top]{subcaption} 
\usepackage{amsmath,graphicx,centernot}
\usepackage{hyperref}
\usepackage{bbm,cancel}
\usepackage{tikz}
\tikzstyle{connection}=[thick,every node/.style={sloped,allow upside down},draw=\ColorFive,opacity=0.9]
\usetikzlibrary{fit}
\usepackage{import}
\usepackage{amsmath,graphicx,centernot}
\usetikzlibrary{arrows,decorations.markings}
\usepackage{enumitem}
\usepackage{comment}

\usetikzlibrary{quotes,arrows.meta}
\usetikzlibrary{positioning}

\usepackage{Box}

\usetikzlibrary{positioning}
\usetikzlibrary{3d} 

\newcommand{\ind}{\mathrel{\perp\mspace{-10mu}\perp}}
\newcommand{\nind}{\centernot{\ind}}
\newcommand{\pa}{{\text{pa}}}
\newcommand{\arr}{-{Triangle[length=2mm, width=2mm]}}

\newcommand{\bx}{\mathbf{x}}

\title{Diagnosing failures of fairness transfer across distribution shift in real-world medical settings}

\author{Jessica Schrouff \thanks{Now at DeepMind}\\\
Google Research \\
\texttt{schrouff@google.com}\\
\And
Natalie Harris \\
Google Research \\
\And
Oluwasanmi Koyejo \\
Google Research \\
\And
Ibrahim Alabdulmohsin \\
Google Research\\
\And
Eva Schnider \thanks{Work performed while interning at Google Research}\\
University of Basel\\
\And
Krista Opsahl-Ong \\
Google Research \\
\And
Alex Brown\\
Google Research \\
\And
Subhrajit Roy \\
Google Research \\
\And
Diana Mincu \\
Google Research \\
\And
Christina Chen \\
Google Research \\
\And
Awa Dieng \\
Google Research\\
\And
Yuan Liu \\
Google Research \\
\And
Vivek Natarajan \\
Google Research \\
\And
Alan Karthikesalingam \\
Google Research \\
\And
Katherine Heller \\
Google Research \\
\And
Silvia Chiappa \\
DeepMind \\
\And
Alexander D'Amour \\
Google Research\\
}

\begin{document}

\maketitle

\begin{abstract}
Diagnosing and mitigating changes in model fairness under distribution shift is an important component of the safe deployment of machine learning in healthcare settings. Importantly, the success of any mitigation strategy strongly depends on the \textit{structure} of the shift. Despite this, there has been little discussion of how to empirically assess the structure of a distribution shift that one is encountering in practice. In this work, we adopt a causal framing to motivate conditional independence tests as a key tool for characterizing distribution shifts. Using our approach in two medical applications, we show that this knowledge can help diagnose failures of fairness transfer, including cases where real-world shifts are more complex than is often assumed in the literature. Based on these results, we discuss potential remedies at each step of the machine learning pipeline.
\end{abstract}

\section{Introduction}
\label{intro}



As progress is made to integrate machine learning (ML) systems in real-world applications, one concern is how the ML model's behaviour might be affected by changes in the data distribution. While extensive work has investigated the effect of distribution shift on ML model performance \citep[see ][ for reviews]{Pan2010-tx,Weiss2016-ch,Farahani2020-dl,Wang2021-fz}, recent work has also highlighted that fairness properties that are satisfied on the data used to develop a model might not always hold if the data distribution changes \citep{Lan2017-ii,kallus2018residual}. In high-risk domains like healthcare, this can translate to `if a model satisfies fairness criteria when trained in ``Hospital A'', it may not satisfy them when tested in ``Hospital B'''. Fueled by this observation, a new field of research is emerging to understand the relationship between ML fairness and robustness to distribution shift \citep[e.g. ][]{Schumann2019-nu,Adragna2020-fs,Veitch2021-rq,Creager2021-ia,Zhang2021-bb,Singh2022-na} and to provide mitigation techniques for obtaining \textit{robustly fair} ML models \citep[e.g. ][]{Singh2021-og,Subbaswamy2020-mi,Slack2020-zz,zhao2020fair,Creager2020}.

Importantly, the success of robustly fair machine learning depends strongly on the nature of the distribution shifts that the system will encounter in practice.
This point has been made elegantly in existing work as a motivation to use causal structure to provide fairness guarantees for specified distribution shifts \citep{Singh2021-og,Subbaswamy2020-mi}. However, to the best of our knowledge, few methods can \textit{diagnose} the nature of the shift and guide towards an appropriate mitigation strategy \cite{Slack2020-zz}.

In this work, we design statistical tests that can, based on a simplified causal graph of an application and a small target dataset, assess the structure of the distribution shift encountered by the system and identify aspects of the data that hinder the transfer of fairness properties. Using applications in dermatology and Electronic Health Records (EHR), we show that this analysis is crucial for selecting an appropriate mitigation strategy. Further, our work also reveals that clinically plausible shifts are often more complex than can be handled by current mitigation techniques, highlighting the need for  additional research into a broader set of remedies across the entire ML pipeline.


\section{Background}
\label{sec:background}
\paragraph{Notation} We consider problems where we are given variables $X$, $A$, $Y$, where $X$ is a set of features, $A$ is one or more sensitive attributes, and $Y$ is an outcome or label to be predicted. Our goal is to build a classification or regression model $f(X')$, where, depending on the context, $X' := (X, A)$ or $X' := X$. We are concerned with the fairness properties of $f$, both on the training distribution and in one or more deployed environments. Depending on the setting, we focus on several statistical definitions of fairness \citep{Barocas2019-kj}: gaps in subgroup accuracy, demographic parity, and equalized odds. See the Supplement for their formulation.

Formally, we represent distribution shifts using the so-called Joint Causal Inference (JCI) framework of \citet{Mooij2020-sv}.
We represent causal relationships between elements of $A$, $X$, $Y$ in a causal Bayesian network (CBN, see the Supplement for an introduction).
In a CBN, a node $U^i$ with an arrow into $U^j$ is called a \textit{direct cause} of $U^j$.
We call all direct causes of a node $U^i$ its causal parents, and denote them $\pa(U^i)$. 
To represent shifts, we augment the CBN with an environment variable $S$, where the event $S=0$ indicates data from the source environment and $S=1$ denotes data from the target environment, so that source data are distributed according to $P(A, X, Y \mid S = 0)$ and the target data to $P(A, X, Y \mid S = 1)$. This is called the \emph{joint causal graph}. In the joint causal graph, arrows from $S$ denote changes to statistical relationships that are induced by the shift. Specifically, an arrow from $S$ to some variable $U$ indicates that the conditional distribution of that variable given its causal parents could change between training and deployment environments, while the absence of such an arrow would indicate that the conditional distribution remains the same, i.e. $P(U \mid \pa(U), S=0) = P(U \mid \pa(U), S=1)$.

\paragraph{Components of Distribution Shifts}
In JCI formalism, the components of a distribution shift are represented by direct arrows from the shift variable $S$ to variables $A$, $Y$, and $X$.
Here, we briefly summarize how these arrows can be interpreted. 
$S \rightarrow A$ arrows represent \emph{demographic shifts}, implying $P(A \mid S=s)$ changes for different values of $s$.
An example of such a shift is a difference in the age distribution of patients in $S=0$ versus $S=1$.
$S \rightarrow X$ arrows represent \emph{covariate shifts}\footnote{In the domain adaptation literature, ``covariate shift'' is often used to mean that \emph{only} $P(X)$ changes. We call this ``exclusive covariate shift''.},
where the conditional distribution of input features $P(X \mid \pa(X), S=s)$ changes for different values of $s$.
For example such a shift could result form using different cameras or views to acquire an image in $S=0$ and $S=1$ (c.f., acquisition shift in \citet{castro2020causality}).
Finally, $S \rightarrow Y$ arrows represent \emph{label shifts}\footnote{As above, we refer to the case where \emph{only} $P(Y)$ changes as ``exclusive label shift''.},
where the conditional distribution of labels $P(Y \mid \pa(Y), S=s)$ changes for different values of $s$.
A label shift can occur through different disease prevalences, or different strategies for obtaining labels, across $S=0$ and $S=1$ (c.f., prevalence shift and annotation shift, respectively, in \citet{castro2020causality}).

In real-world applications, distribution shifts often include multiple components, where there are multiple arrows from $S$ to $\{A, X, Y\}$.
We call such shifts \emph{compound}.
Compound shifts often occur when there is an unobserved variable that affects multiple variables whose distribution changes between source and target.
An example is that of a new device on the market \citep{Finlayson2021-zo}. Such a device could create qualitatively different images $X$, but it could also change the observed patient population by being made more available to certain subsets of demographic groups that tend to be more or less healthy.
In this case, the shift induced by introducing the device simultaneously affects $A, X$ and $Y$.
\vspace{-0.3cm}

\paragraph{Shift Structure and Fairness Transfer}
The transfer of fairness depends on the causal structure of the application and the structure of the shift \citep{Singh2021-og, Singh2022-na}.
In particular, implications for fairness transfer can be directly read off of conditional independences with respect to the shift $S$ in the joint causal graph.  For example, for classifiers that take a set of variables $\{\mathbf V\}$ as input, the following metrics are known to be invariant across the shift under the following conditional independence conditions:
\begin{itemize}[noitemsep,nolistsep]
\item Classification error ($P(Y \neq f(\mathbf V) \mid \mathbf V, S)$) is invariant if $Y \ind S \,|\, \{\mathbf{V}\}$ \citep{Magliacane2018-oh,Singh2021-og}.
\item Demographic parity gap ($\max_{a \in \mathcal A}E[f(\mathbf V) \mid S, A=a] - \min_{a \in \mathcal A}E[f(\mathbf V) \mid S, A=a]$) is invariant if $\{\mathbf{V}\} \ind S \,|\, A$ \citep{Singh2021-og}.
\item Equality of odds gap ($\max_{a,y \in \mathcal{A} x \mathcal Y}E[f(\mathbf V) \mid S, A=a, Y=y] - \min_{a,y \in \mathcal{A} x \mathcal Y}E[f(\mathbf V) \mid S, A=a, Y=y]$) is invariant if $\{\mathbf{V}\} \ind S \,|\, Y, A$ \citep{Singh2021-og}.
\end{itemize}

In JCI formalism, the above conditions imply that fairness transfer requires certain arrows from $S$ to $\{A, X, Y\}$ to be missing.
The nature of the shift also strongly influences which mitigation strategies are applicable.
We discuss implications for mitigation strategies in more detail in Sec.\ref{sec:rel_works}.
\vspace{-0.3cm}

\section{Empirical tests for assessing shift structure}
\label{sec:shift_detection}
\begin{algorithm}[!t]
\caption{(Conditional) independence testing assessing the nature of shift $S$ on a single variable $U \in \mathcal{G}$. $U$ represents the feature values or its summary if high-dimensional. $\odot$ is the Hadamard product.}
\label{alg}
\begin{algorithmic}
\Require source dataset $\mathcal{D}$, target dataset $\mathcal{D'}$, a joint causal graph $\mathcal{G}$
\State Split $\mathcal{D}$ into $\mathcal{D}_{w}$, $\mathcal{D}_{t}$; Split $\mathcal{D'}$ into $\mathcal{D'}_{w}$, $\mathcal{D'}_{t}$
\For{1:$n_{bootstrap}$} 
    \State  Sample $\mathcal{V}$ from $\mathcal{D}_{t}^{n_0\times l}$ and $\mathcal{V'}$ from $\mathcal{D'}_{t}^{n_1\times l}$
    \State Set $w_s = 1^{n_s}$ for $s \in \{0,1\}$.
   \State  \If{$\pa(U) \hspace{1mm} \texttt{not} \hspace{1mm} \emptyset$} \Comment{Compute balancing weights}
        \State  Sample dataset ${\cal Q}=\{\pa(U) |S=0\}$ from $\mathcal{D}_w$ and ${\cal Q'}=\{\pa(U)|S=1\}$ from $\mathcal{D'}_w$
        \State  Build predictor $P(S | \pa(U))$ classifying ${\cal Q}$ from ${\cal Q'}$
        \State $w_s(\pa(U)) \propto P(\pa(U) \mid S=s)^{-1} \in \mathcal{R}^{n_s}$ for $s \in \{0,1\}$.
        \EndIf
    \State Compute t-statistic between $[w_0 \odot \mathcal{V}(U); -w_1 \odot \mathcal{V'}(U)]$ and 0
\EndFor
\State Compute p-value from t-test of bootstrapped t-statistics against 0
\State \textbf{return} p-value\Comment{The p-value that $U \ind S \mid \pa(U)$}
\end{algorithmic}
\end{algorithm}

Given the central role of distribution shift structure in fairness transfer, our main methodological proposal is to investigate the structure of shifts empirically when fairness fails to transfer.
In this section, we outline a general strategy for testing which components of a shift are present, given data from source and target distributions, and a BCN for the application.

\noindent \textbf{Approach. }
For each variable $U$ in the graph (i.e. each $A$, $X$ and $Y$), we assess whether there is a direct effect of the environment $S$ on $U$.
We test the null hypothesis $H_0$: $P(U \mid \pa(U), S=0) = P(U \mid \pa(U), S=1)$.
Rejecting this hypothesis implies an arrow $S \rightarrow U$ in the joint causal graph.
If the marginal distribution of $U$ differs across source and target distributions, this comparison can be challenging.
To isolate the direct effect of $S$,
we compute weights that balance the causal parents $\pa(U)$ across environments \citep{Rosenbaum1983-zd,Imai2004-uh}.
We then test the hypothesis that the reweighted distributions of $U$ match across the environments $S$.
For simplicity, we test whether the reweighted means are equal, and in cases where the dimensionality of $U$ is high, we conduct this test on functions of $U$ that reduce the dimensionality, following strategies in \citet{Rabanser2019-gu}.
\footnote{A rejection of these simplified hypotheses implies that the original null hypothesis should also be rejected.}
This procedure is described in Algorithm~\ref{alg} and detailed in the Supplement.

\noindent \textbf{Validation. } We conduct three validation experiments for our testing procedure using data from our dermatology application in the Supplement. We first assess Type I error in an experiment where we test random splits of the same data against each other. Our test displays a false positive rate similar to our threshold of $5\%$ for hypothesis testing. We however note that the variance of this result increases with smaller numbers of samples. We then confirm non-trivial power in an experiment where we introduce a shift by subsampling younger patients with a particular skin condition in the source dataset (a shift on $A$ and $Y$). Our testing approach correctly identifies the dimensions of $A$ and $Y$ that were affected by $S$ due to the engineered shift. We replicate this test for a shift in $Y$ only, with the same result.

\section{Distribution shifts in real-world healthcare applications}
\label{sec:problem_setting}
Here, we present two case studies from the healthcare domain in dermatology and in clinical risk prediction using Electronic Health Records in which fairness does not transfer (Fig. \ref{fig:no_transfer}).
As per \citep{Koh2020-cr}, we consider the setting of `dataset shift', whereby a model is developed on the source data and tested on the target data\footnote{We discuss fine-tuning and joint training in Sec.\ref{sec:rel_works}, but this is not the setup of our work}, which is a common setting in medical applications \citep{Wiens2014-mu,Zech2018-os,Koh2020-cr}. 

\subsection{Predicting skin conditions in dermatology}
\label{sec:derm}

In this task, we predict common skin conditions (26 conditions plus an additional `other' category to capture the long tail), based on one or multiple images of the pathology of interest and additional metadata \citep{Liu2020-qr}. The data used as source is a subset of the data described in \citet{Liu2020-qr}. Briefly, it consists of de-identified retrospective adult cases from a teledermatology service serving 17 sites from 2 states in the United States. Each case contained 1-6 clinical photographs of the affected skin areas as well as patient demographic information and medical history (see Table S1 of \citep{Liu2020-qr}). We split the source data such that it comprises 12,024 cases for training, 1,925 for validation and hyper-parameter tuning and 1,924 for testing.

We consider how the fairness properties of a model trained in this setting transfer to a target dataset, unobserved at model development time, consisting of 1,843 labelled teledermatology cases from Colombia. In the Supplement, we consider transfer to a second target from skin cancer clinics in Australia, as well as the properties of a model jointly trained on the source and the second target, and deployed on the first target data. The model we consider is a ResNet architecture, as in \citep{Liu2020-qr,Roy2021-xp}.

\begin{figure*}[!t]
\centering  
\begin{subfigure}[!t]{0.3\textwidth}
(a) Dermatology
\end{subfigure} 
\begin{subfigure}[!t]{0.6\textwidth}
\includegraphics[width=\linewidth,valign=t]{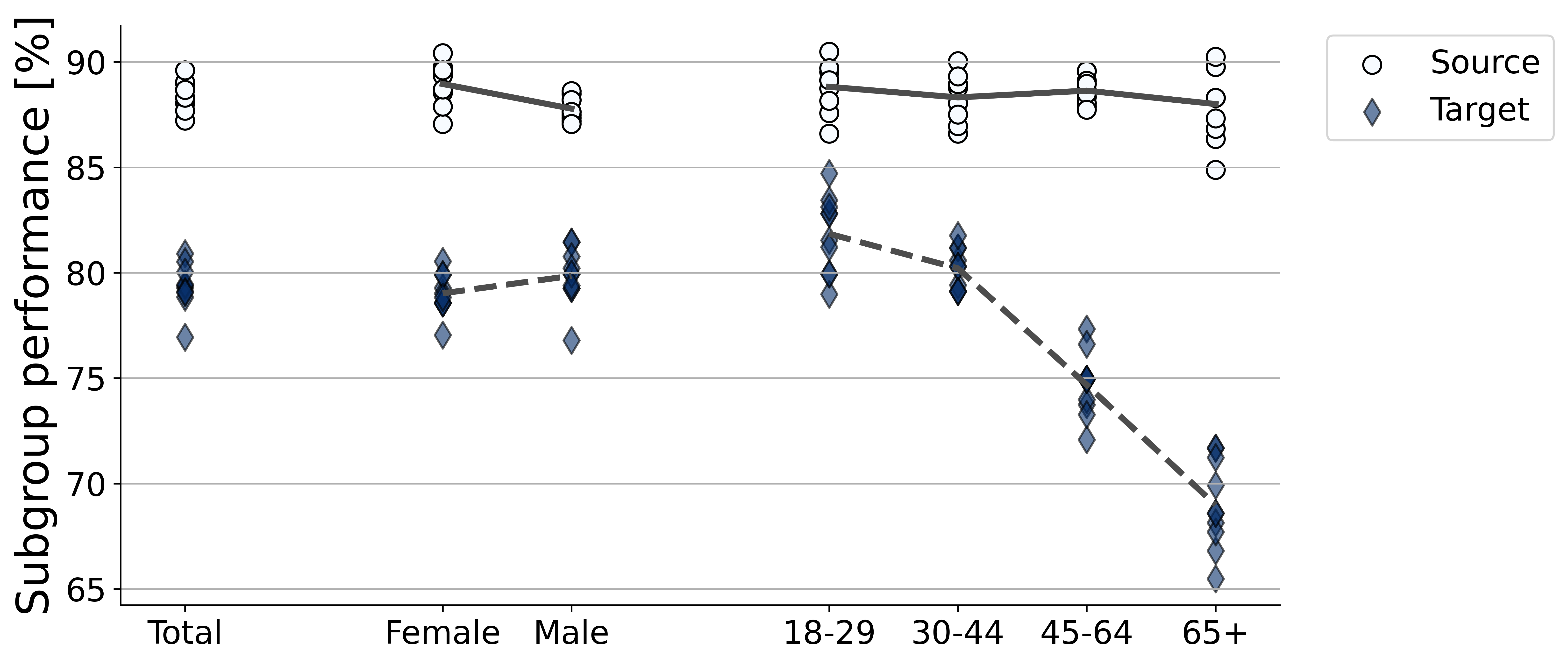} 
\end{subfigure} \\
\begin{subfigure}[!t]{0.3\textwidth}
(b) EHR
\end{subfigure} 
\begin{subfigure}[!t]{0.6\textwidth}
\includegraphics[width=\linewidth,valign=t]{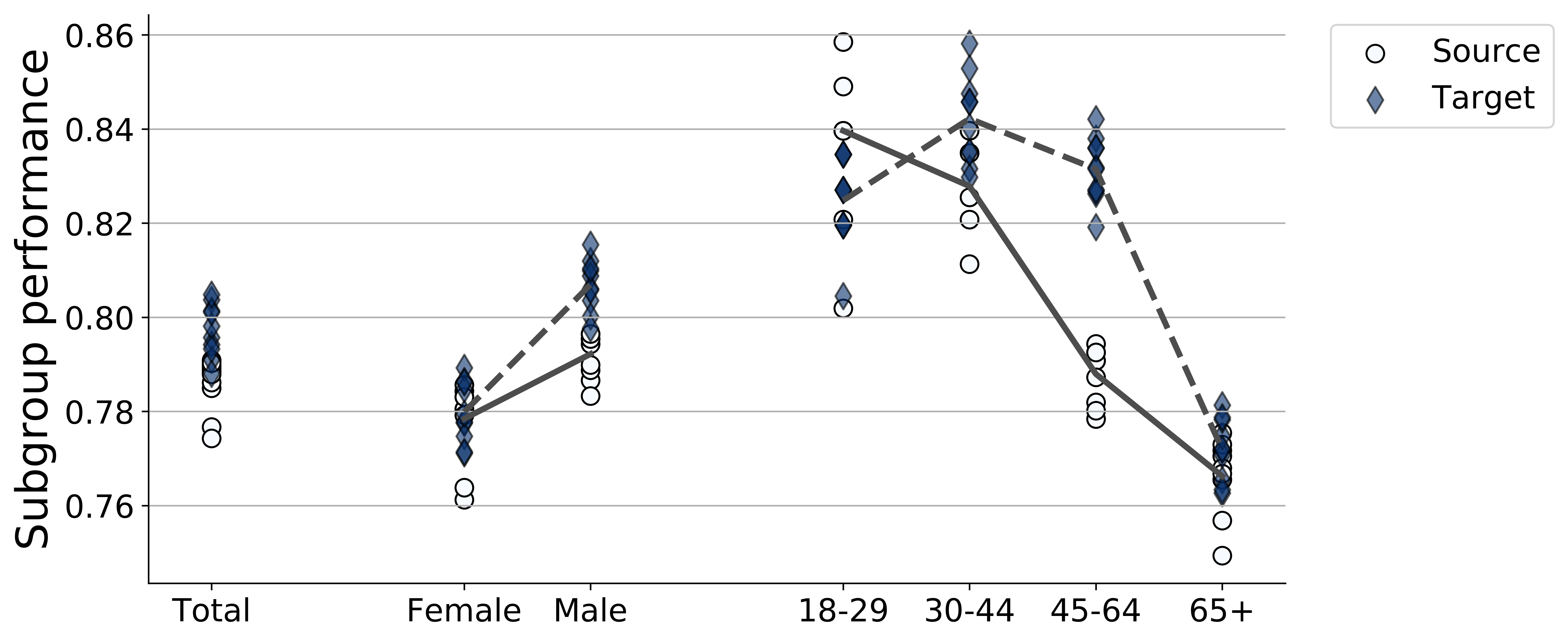} 
\end{subfigure}
\caption{Model performance across subgroups (age and sex) on the source (circles with plain line) and target (diamonds with dashed line). Each marker represents one replicate of the model. (a) Top-3 accuracy (in \%) in the dermatology application. (b) Accuracy in EHR.}
\label{fig:no_transfer}       
\end{figure*}

We assess model performance and fairness across 10 replicates of the model on the source and on the target (Fig.~\ref{fig:no_transfer}). On the source data, the model performs similarly across all subgroups and could reasonably be assessed as `fair'.
However, the maximum gap between age groups increases starkly on the target (from $1.05\%$ to $12.86\%$).

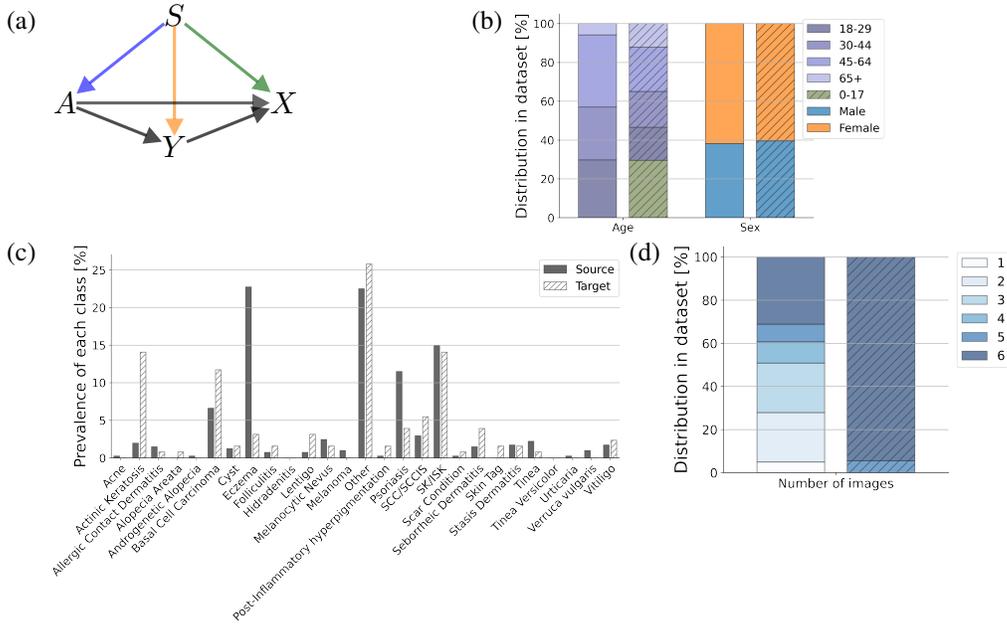
\begin{figure*}
\begin{subfigure}[t]{0.03\textwidth}
(a)
\end{subfigure}
\begin{subfigure}[t]{0.4\textwidth}
\resizebox{0.6\linewidth}{!}{
\begin{tikzpicture}[baseline={([yshift={-\ht\strutbox}]current bounding box.north)},outer sep=0pt,inner sep=0pt]
\node (S) at (-0.25,-3) {$S$};
\node (X) at (1,-4) {$X$};
\node (A) at (-1.5,-4) {$A$};
\node (Y) at (-0.25,-4.5) {$Y$};
\draw[line width=1pt,orange, \arr, opacity=0.6](S)--(Y);
\draw[line width=1pt,blue, \arr, opacity=0.6](S)--(A);
\draw[line width=1pt,green!40!black, \arr, opacity=0.6](S)--(X);
\draw[line width=1pt,black, \arr, opacity=0.6](A)--(X);
\draw[line width=1pt,black, \arr, opacity=0.7](A)--(Y);
\draw[line width=1pt,black, \arr, opacity=0.7](Y)--(X);
\end{tikzpicture}}
\end{subfigure}
\begin{subfigure}[t]{0.03\textwidth}
(b)
\end{subfigure}
\begin{subfigure}[t]{0.45\textwidth}
\includegraphics[width=0.8\linewidth,valign=t]{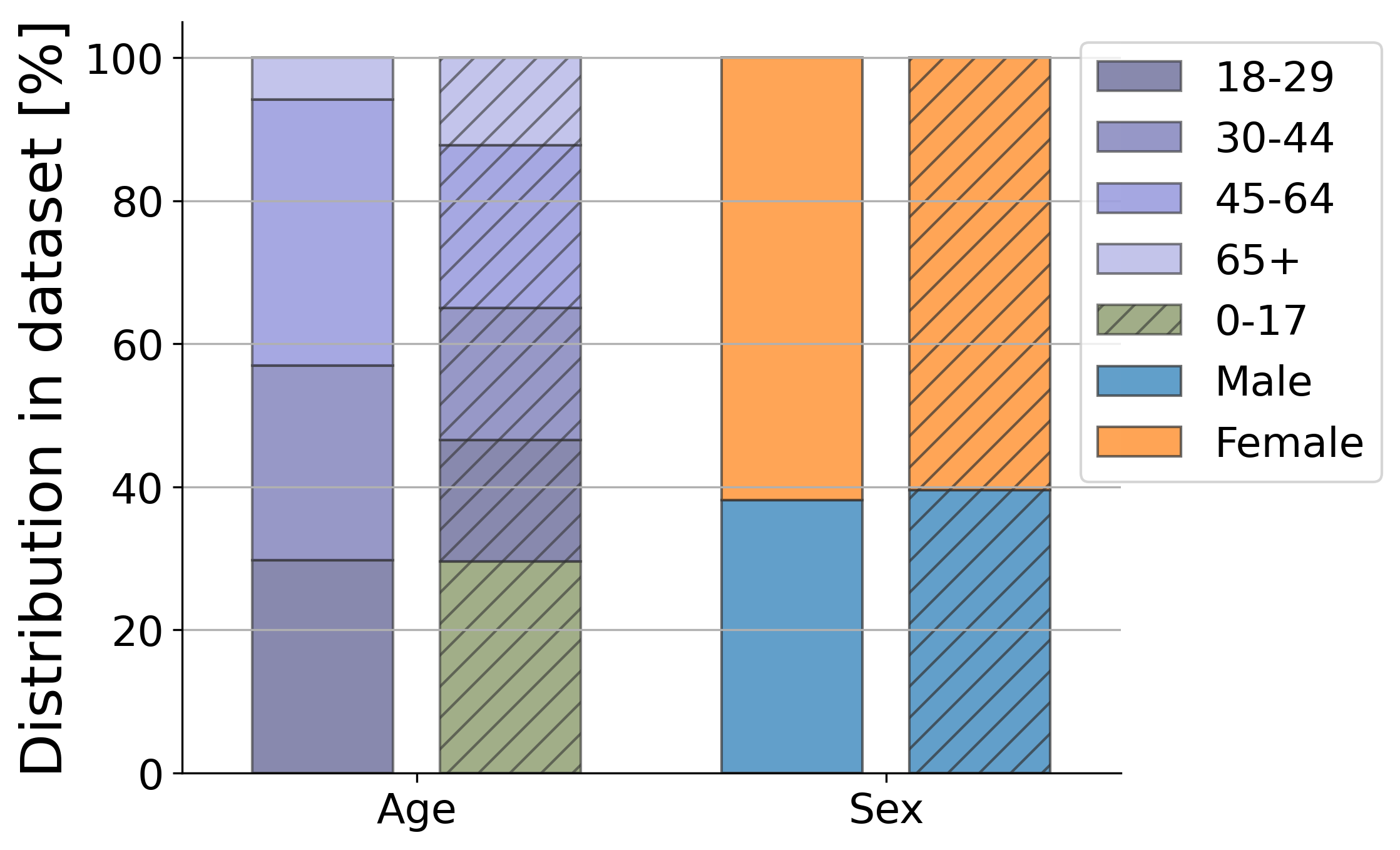} 
\end{subfigure}\\
\begin{subfigure}[t]{0.03\textwidth}
(c)
\end{subfigure}
\begin{subfigure}[t]{0.55\textwidth}
\includegraphics[width=\linewidth,valign=t]{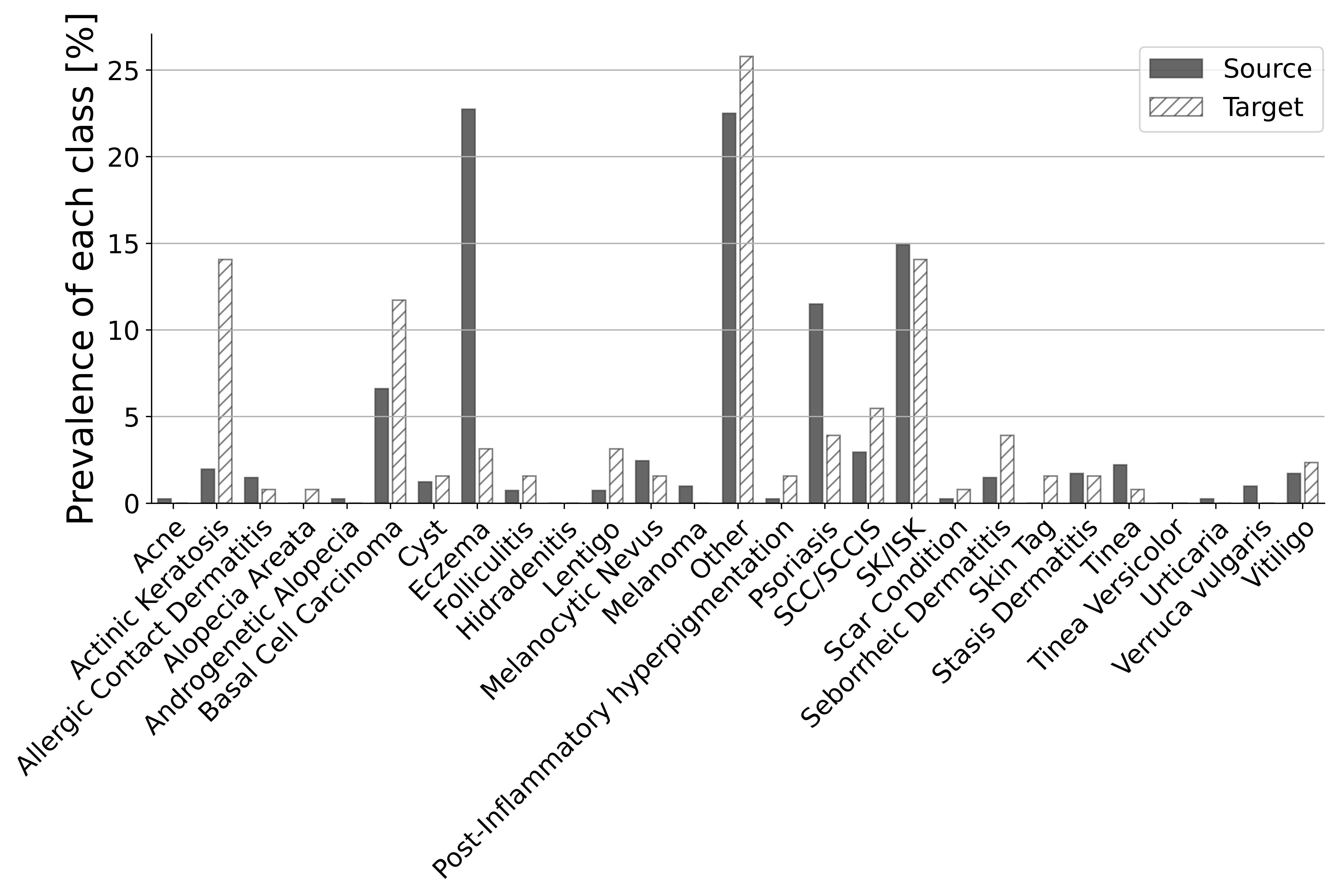}
\end{subfigure}
\begin{subfigure}[t]{0.03\textwidth}
(d)
\end{subfigure}
\begin{subfigure}[t]{0.33\textwidth}
\includegraphics[width=\linewidth,valign=t]{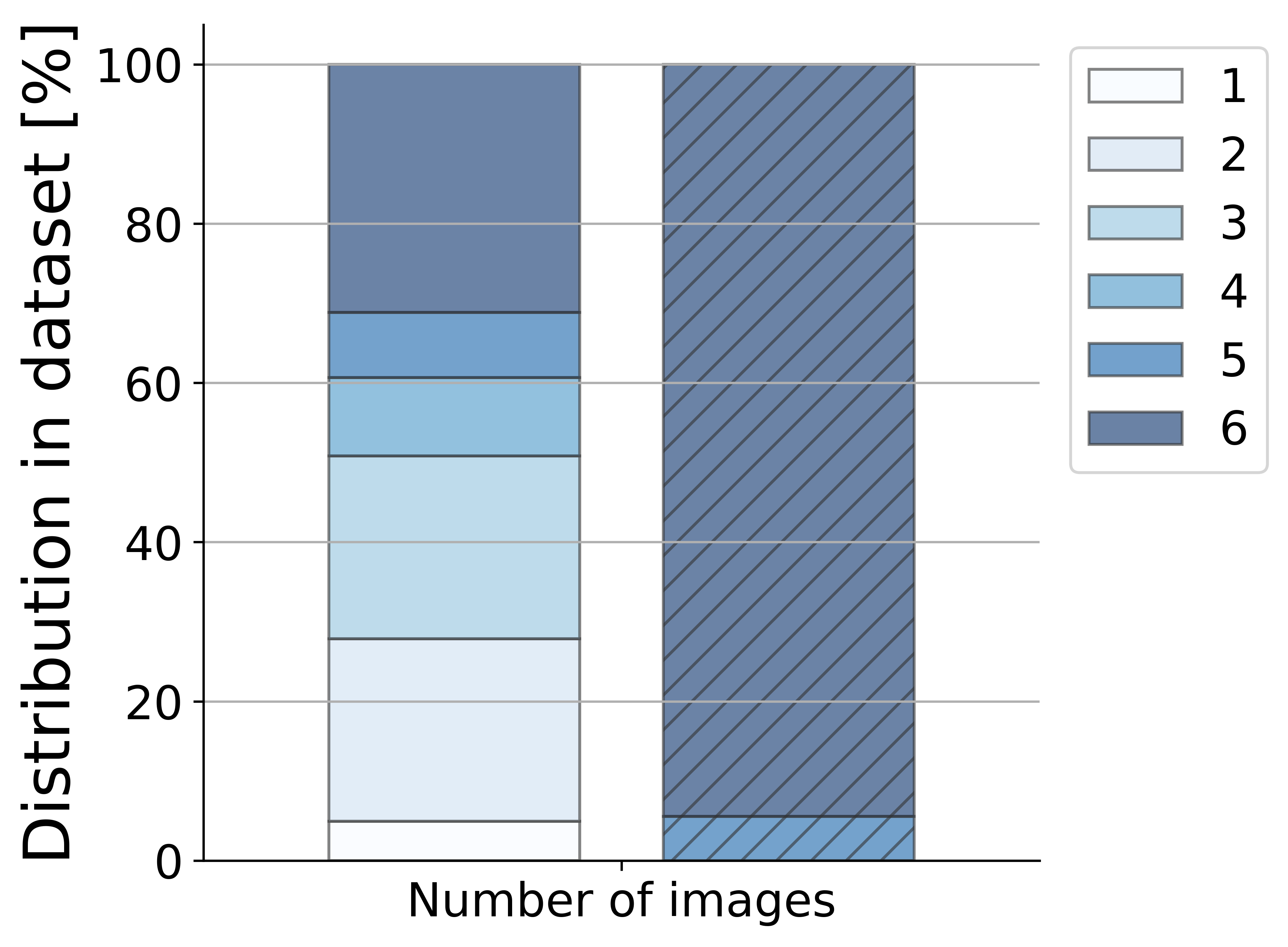}
\end{subfigure}
\caption{A compound shift in dermatology. (a) Simplified causal structure of the application (see text for variable description). Colored arrows represent statistical dependence as identified by Algorithm~\ref{alg}. (b) Distribution in the source and in the target dataset of the sensitive attributes, computed in terms of percentage of pre-defined subgroups. (c) Prevalence of each condition for female patients over 65 years of age. (d) Distribution of the number of images in cases labelled as `SK/ISK' in older females. Each case includes min. 1, max. 6 images whose embeddings are then averaged. The distributions in the source data are presented on the left, and their corresponding distributions in the target data on their immediate right with a hashed pattern.}
\label{fig:derm_shift2}       
\end{figure*}

To understand the failure of fairness transfer in this context, we investigate the structure of the shift by referring to Algorithm~\ref{alg}.
We begin by drawing a simplified causal graph $\mathcal{G}$ of the dermatology task at hand (Fig.\ref{fig:derm_shift2}(a)). We consider this an anti-causal learning task \cite{scholkopf2012on}, in which the skin condition label $Y$ is a cause of the input image $X$, and consider demographic information $A$ to be a causal parent of both $X$ and $Y$.
We omit variables that are unobserved in either the source or target data.
\footnote{The metadata available in the source data includes co-morbidities/medical history and symptoms, but these are absent in the target datasets.}

We now test whether the shift $S$ has direct effects on $A$, $Y$, or $X$.
Overall, we find that the shift is compound, and affects all three aspects of the data distribution.
We begin with $A$, which comprises age and sex.\footnote{Sex is mainly recorded by clinicians in the source. Sex is self-reported in the target.}
Because $A$ has no causal parents in our graph (besides, potentially, $S$) we directly assess differences in age and sex across environments using classical tests.
Age has a different distribution across the two datasets (see Fig.~\ref{fig:derm_shift2}(b)), with the population in the source data being typically younger (median: 40 years old, 25$\%$ quantile: 27, 75$\%$ quantile: 54) than in the target data (median: 44 years old, 25$\%$ quantile: 30, 75$\%$ quantile: 59).
Sex distributions are well matched across the source and target.
These results suggest a direct effect of $S$ on $A$, and we add this relationship to the causal graph (Fig.~\ref{fig:derm_shift2}(a), in purple).

We then assess whether $S$ directly affects the labels $Y$.
In our graph, $A$ is a causal parent of $Y$, so following our testing strategy in Algorithm~\ref{alg}, we test whether $P(Y \mid A, S=0) = P(Y \mid A, S=1)$.
(Recall that conditioning on $A$ isolates the direct effect of $S$ on $Y$.)
Using 27 univariate tests (one per condition, \citep{Rabanser2019-gu}), we obtain significant differences between $P(Y \mid A, S=0)$ and $P(Y \mid A, S=1)$ for 3 conditions ($p<0.05$, Bonferroni corrected). Fig.\ref{fig:derm_shift2}(c) illustrates the label shift in a specific age and sex subgroup (here females aged 65+, 409 cases in the source, 128 in the target data). We see that the source data includes more cases of `eczema' and `psoriasis' while conditions like `actinic keratosis' and `seborrheic dermatitis' are more represented in the target.
These results suggest that the environment also directly affects the labels (orange link in Fig.\ref{fig:derm_shift2}(a)). 

Finally, we consider whether the features of the images themselves (designated by $X$) are directly affected by $S$.
Conditioning on causal parents of $X$, we test whether $P(X \mid Y, A, S=0)=P(X \mid Y, A, S=1)$.
We summarize $X$ by training a model $f(X) \rightarrow Y$ on the source, a strategy proposed in \citet{Rabanser2019-gu}. Our weighted tests suggest a significant difference between the two distributions ($p<0.05$ corrected on 1 dimension). 
Fig.\ref{fig:derm_shift2}(d) illustrates this difference by computing the number of images per case in the group of older females considered above with cases labelled as `SK/ISK' ( source median: 3, $n=61$, target median: 5, $n=18$). This result suggests the existence of the direct path $S \rightarrow X$ and we add this relationship to the graph (green link in Fig.\ref{fig:derm_shift2}(a)).

Based on these results, we observe that all aspects of the data are affected by the environment. Thus, none of the conditional independence conditions for fairness transfer hold in this setting.
However, we also see that some conditions might display invariance once controlled for demographics, which suggests possible mitigations that we explore in Section~\ref{sec:rel_works}.
The Supplement includes experimental details and further results on the second target and the joint training.

\subsection{Clinical risk prediction from Electronic Health Records}
\label{sec:ehr}
In this application, we predict the clinical outcomes of patients based on EHRs. EHRs record the time series of interactions between a patient and the clinical system (e.g. medication, labs, vitals,~\dots).
Distribution shift is a major concern for systems that consume EHR data, which can be heterogeneous across centers \citep{Wilson2021-uq, Singh2022-na}, and even within a single system \citep{Nestor2019-dp}.
As an example, over half of hospitals in the US with intensive care units (ICUs) only have a single unit where all critically ill patients (medical, surgical, cardiac, etc.) are admitted \footnote{Critical Care Statistics. \url{https://www.sccm.org/Communications/Critical-Care-Statistics}}. Risk scores, such as the Acute Physiology and Chronic Health Evaluation II (APACHE II) score are used to evaluate all critically ill patients, but do not perform equally across clinical subgroups, e.g. underperforming on patients with cardiac disease~\citep{Pierpont1999-qb}. It would be desirable to avoid such issues with risk scores constructed through machine learning.
We use the open access, de-identified Medical Information Mart for Intensive Care III (MIMIC-III) dataset \citep{Johnson2016-eu}, which consists of data from admissions to ICUs at the Beth Israel Deaconess Medical Center between 2001 and 2012. This clinical system has the benefit of having separate specialized ICUs and allows to assess the generalizability of risk scores estimated in one clinical population to another. Based on clinical input, we consider the Medical ICU (MICU), Surgical ICU (SICU) and Trauma Surgical ICU (TSICU) to be generalist ICUs (source data); whereas the Cardiac Surgery Recovery Unit (CSRU) and Coronary Care Unit (CCU) are specialized ICUs (target data). This split leads to 17,641 patients included in the source dataset and 10,442 patients in the target dataset, after selecting adult patients with a length of stay of minimum one day and with a recorded care unit. Our goal is to obtain a robustly fair model that predicts prolonged ICU stay (i.e. length of stay $>3$ days, as in \citep{Wang2019-ac,Pfohl2021-xn}) using data from the first 24 hours of first ICU admission and a recurrent architecture \citep{Tomasev2019-tu,Tomasev2021-uf,Roy2021-rr}. The model is trained on $80\%$ of the source data, tuned using a separate split of $10\%$ and tested on the remaining $10\%$. See the Supplement for further details. Noteworthy, the model does not have access to the `reason for visit'.

Figure~\ref{fig:no_transfer}(b) shows that the fairness properties of this model, as measured by gaps in subgroup performance, demographic parity (DP) and equalized odds (EO), do not transfer across ICU types.
The model displays unfairness w.r.t. age in both datasets (maximum gap $\sim7\%$, similar DP and EO scores). However, the pattern of unfairness is different in the two datasets, with e.g. the 45-64 years old subgroup being under-served in the source but over-served in the target. The performance gap between sex groups increases from $1\%$ in the source (DP: $0.002 \pm 0.002$) to $2.7\%$ (DP: $0.016 \pm 0.003$)
on the target.

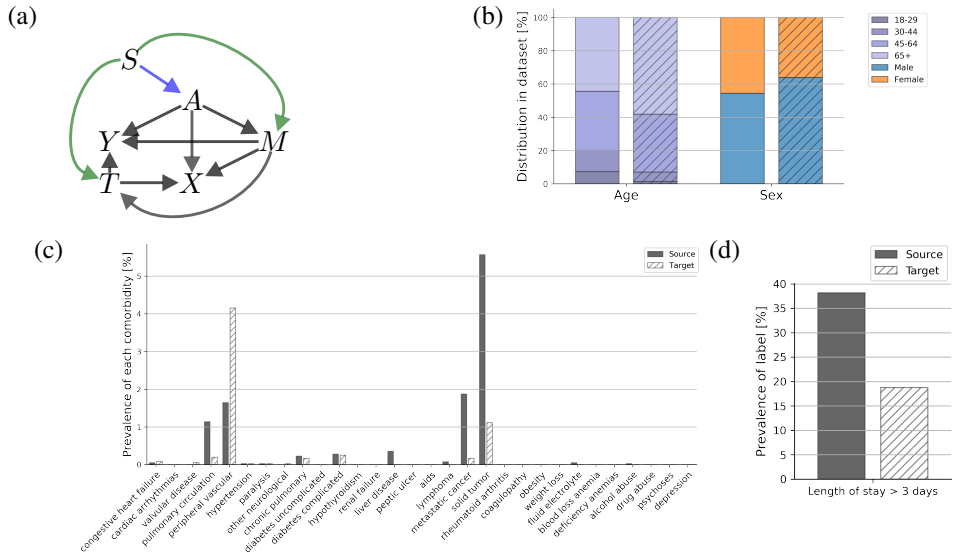
\begin{figure*}[!t]
\centering
\begin{subfigure}[t]{0.03\textwidth}
(a)
\end{subfigure}
\begin{subfigure}[t]{0.4\textwidth}
\resizebox{0.6\linewidth}{!}{
\begin{tikzpicture}[baseline={([yshift={-\ht\strutbox}]current bounding box.north)},outer sep=0pt,inner sep=0pt]
\node (S) at (0.25,-0.5) {$S$};
\node (X) at (1,-2) {$X$};
\node (A) at (1,-1) {$A$};
\node (M) at (2,-1.5) {$M$};
\node (Y) at (0,-1.5) {$Y$};
\node (T) at (0,-2) {$T$};
\draw[line width=1pt,blue, \arr, opacity=0.6](S)--(A);
\draw[line width=1pt,green!40!black, \arr, opacity=0.6](S) to [bend left=+78] (M);
\draw[line width=1pt,green!40!black, \arr, opacity=0.6](S) to [bend right=+78] (T);
\draw[line width=1pt,black, \arr, opacity=0.6](A)--(X);
\draw[line width=1pt,black, \arr, opacity=0.7](A)--(M);
\draw[line width=1pt,black, \arr, opacity=0.7](A)--(Y);
\draw[line width=1pt,black, \arr, opacity=0.7](M)--(Y);
\draw[line width=1pt,black, \arr, opacity=0.7](M)--(X);
\draw[line width=1pt,black, \arr, opacity=0.6](M) to [bend left=+58] (T);
\draw[line width=1pt,black, \arr, opacity=0.7](T)--(X);
\draw[line width=1pt,black, \arr, opacity=0.7](T)--(Y);
\end{tikzpicture}}
\end{subfigure}
\begin{subfigure}[t]{0.03\textwidth}
(b)
\end{subfigure}
\begin{subfigure}[t]{0.5\textwidth}
\includegraphics[width=0.8\linewidth,valign=t]{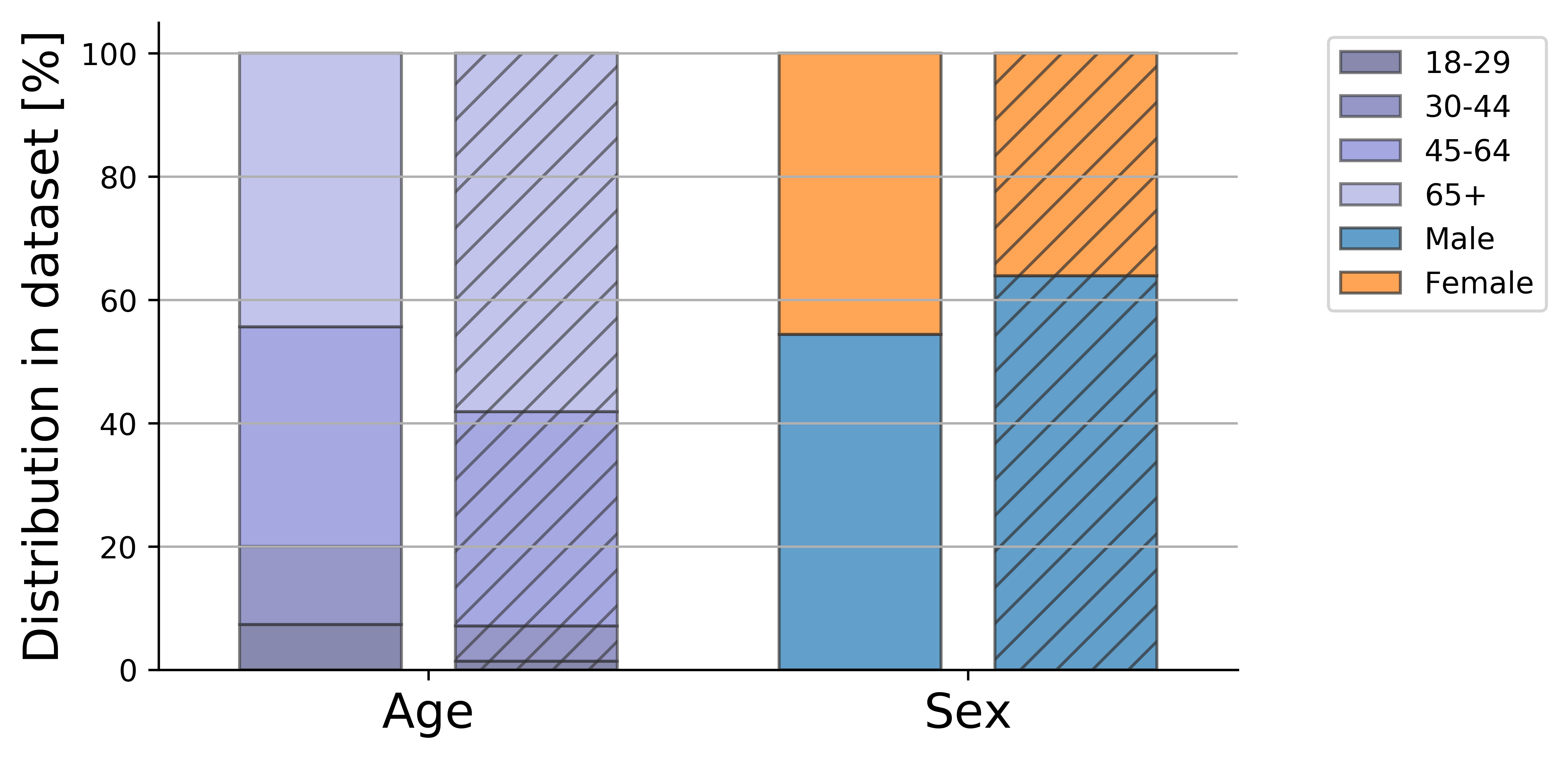} 
\end{subfigure}\\
\begin{subfigure}[t]{0.03\textwidth}
(c)
\end{subfigure}
\begin{subfigure}[t]{0.6\textwidth}
\includegraphics[width=\linewidth,valign=t]{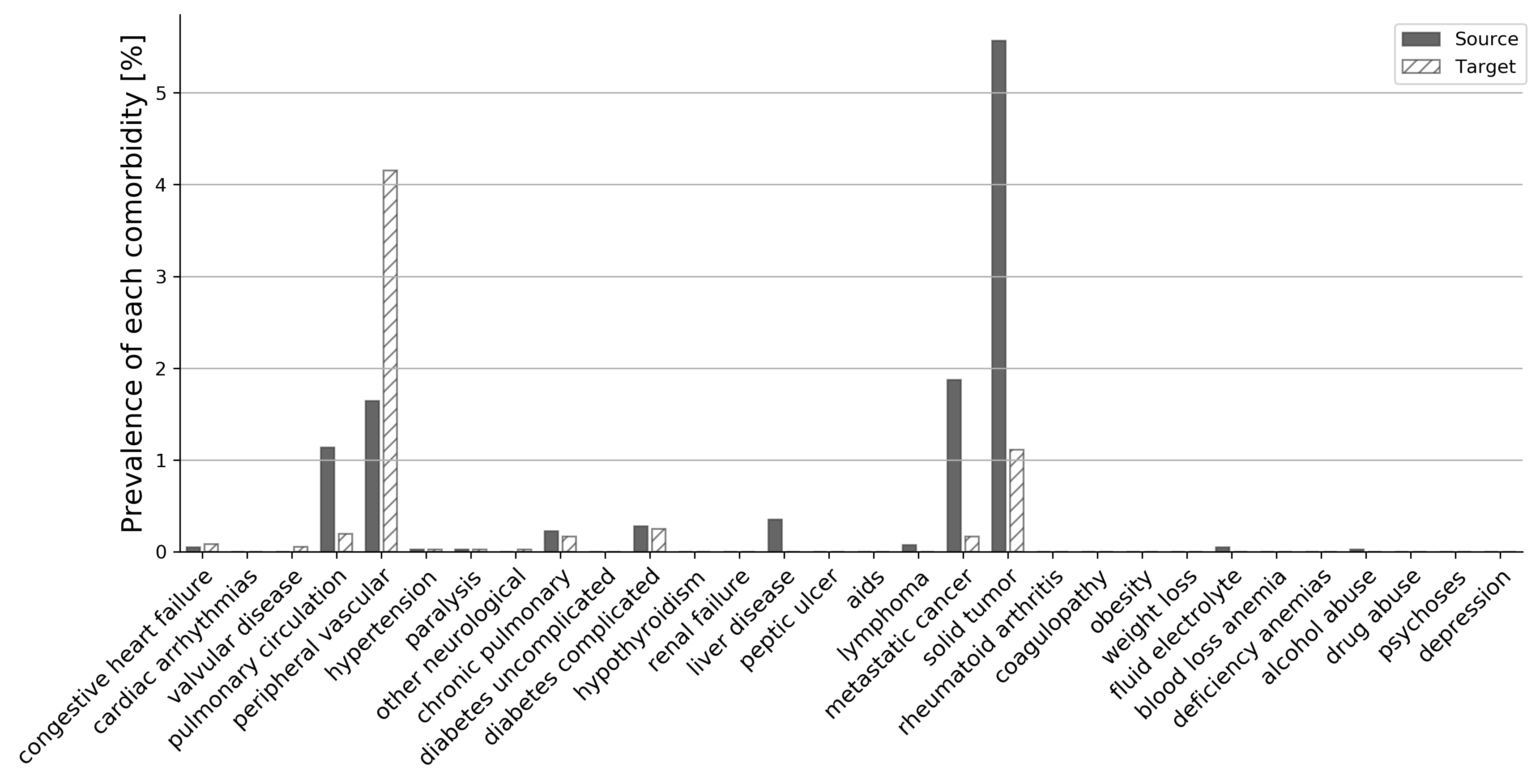}
\end{subfigure}
\begin{subfigure}[t]{0.03\textwidth}
(d)
\end{subfigure}
\begin{subfigure}[t]{0.25\textwidth}
\includegraphics[width=0.8\linewidth,valign=t]{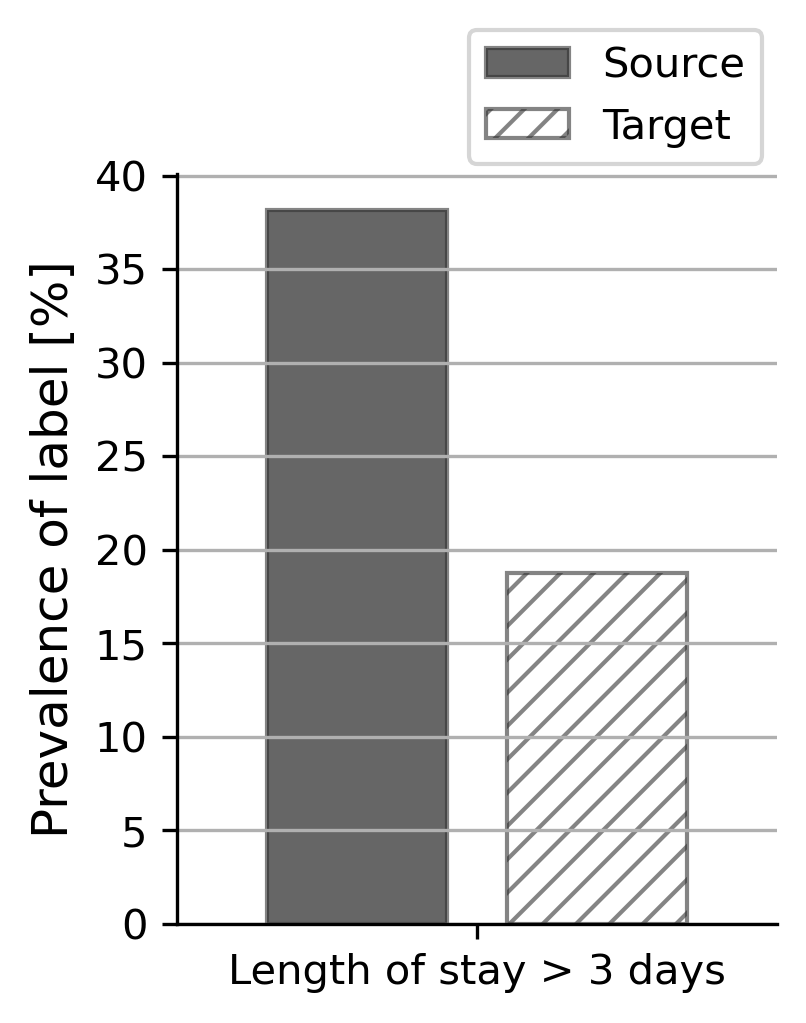}
\end{subfigure}
\caption{A compound shift in EHR. (a) Causal graph for this application (see text for variable description). Colored arrows represent statistical dependence as identified by Algorithm~\ref{alg}. (b) Distribution of the sensitive attributes, computed in terms of percentage of pre-defined subgroups. (c) Prevalence of comorbidities for male patients over 65 years of age. (d) Prevalence of the label for male patients over 65 with a solid tumor comorbidity and under vasopressors/inotropes. The distributions in the source are represented on the left, and their corresponding distributions in the target on their immediate right with a hashed pattern.}
\label{fig:ehr_shift}       
\end{figure*}

To understand this failure of fairness transfer, we examine the structure of the shift.
First, we draw a causal graph for the variables in this problem, building on the work in \citet{Singh2021-og}.
Here, we incorporate demographics $A$ as age and self-reported sex, comorbidities $M$ as previous medical history defined by ICD-9 codes \citep{Charlson1987-xb,Deyo1992-ct,Halfon2002-nq,Li2008-go}, observed features $X$ as all labs and vitals, and $Y$ as our prolonged length of stay label. As a more comprehensive feature set leads to better predictive performance \citep{Roy2021-rr}, we use all 59,351 features at our disposal and add `treatments' $T$ (e.g. medications) to the graph (Fig.~\ref{fig:ehr_shift}(a)). In our graph, $S$ represents the ICU unit type that a patient is admitted to.

We now test whether the shift $S$ has direct effects on the variables in this problem using our proposed approach.
Beginning with $A$, which has no causal parents, we test whether $P(A \mid S=0) = P(A \mid S=1)$.
We estimate the distribution of age and sex in each dataset (Fig.~\ref{fig:ehr_shift}(b)). We observe that the population in the source is typically younger than the population in the target (t-test: $p<0.0001$). In addition, the proportion of males in the source and target differ (source: 50\%, target: 65\%). Therefore, $S \rightarrow A$.

To assess the shift's effect on comorbidities $M$, we refer to Elixhauser and Vanwalraven scores \citep{Elixhauser1998-ra, Van_Walraven2009-ei} computed on the source and target data \citep{Johnson2018-zj}\footnote{Code publicly available at \url{https://doi.org/10.5281/zenodo.821872}}. Figure~\ref{fig:ehr_shift}(c) displays the distribution of the 30 comorbidities for older males ($n=3,951$ in source and $n=3,588$ in target). As expected, even after adjusting for the causal parent $A$, patients in the target data have a higher prevalence of comorbidities related to the peripheral vascular system than in the source data and $M \nind S \,|\, A$ for 5 comorbidities (weighted tests: $p<0.05$, Bonferroni corrected).

We repeat the analysis to assess the direct effect of $S$ on the treatments ($P(T \,|\, A, M, S=0) = P(T \,|\, A, M, S=1): p<0.05$), on the labs and vitals ($P(X \,|\, A, M, T, S=0) = P(X \,|\, A, M, T, S=1): p>0.05$) and on the label ($P(Y | A, M, T, S=0) = P(Y \,|\, A, M, T, S=1): p>0.05$) in the Supplement. Figure~\ref{fig:ehr_shift}(d) displays the prevalence of prolonged length of stay for older males with a solid tumor comorbidity and receiving vasopressors/inotropes ($n=55$ in source and $n=16$ in target).

Our results suggest that the shift is compound, i.e. the ICU unit a patient is admitted to changes the relationships between multiple variables in our causal graph. We note that this conclusion is not surprising clinically, given that the environment $S$ is correlated with the reason for their ICU admission, and that this reason for admission (unobserved) is a main driver for all variables. As in the dermatology case, the compound shift leads to multiple paths for $S$ and $A$ to affect the covariates $M$ and $X$ and the label $Y$, explaining the failure of fairness transfer.

\section{Mitigation and related work} 
\label{sec:rel_works}


A variety of fairness, robustness and robustly fair mitigation strategies have been proposed. We contextualize a number of methods in terms of anti-causal and causal learning tasks \citep{scholkopf2012on} in the Supplement, and highlight how understanding the nature of the distribution shift can guide mitigation.

\noindent \textbf{Fairness mitigation. } Pre-, in- and post-processing techniques for mitigating unfairness have been proposed \citep[see ][ for overviews]{Barocas2019-kj,oneto2020fairness}. Fairness metrics and mitigation techniques currently used in real-world applications are typically evaluated on a single learning task or data distribution \citep[e.g.][]{Zhao2017-zq,agarwal18a,Wang2020-bb,Pfohl2021-xn,Cherepanova2021-qh}. More similar to our settings, the work by \citet{Creager2020} addresses fairness in dynamical systems, although only for low-dimensionality variables. 
Based on our causal framing, applying fairness mitigations in the source domain could potentially aid in fairness transfer by cutting some edges in the underlying CBN \citep{Veitch2021-rq}.
However, fairness mitigation on the source only leads to provable fairness transfer in specific settings, i.e. mostly when the shift is demographic (Supplement).

\begin{figure*}[t]
\centering
\includegraphics[width=0.6\linewidth]{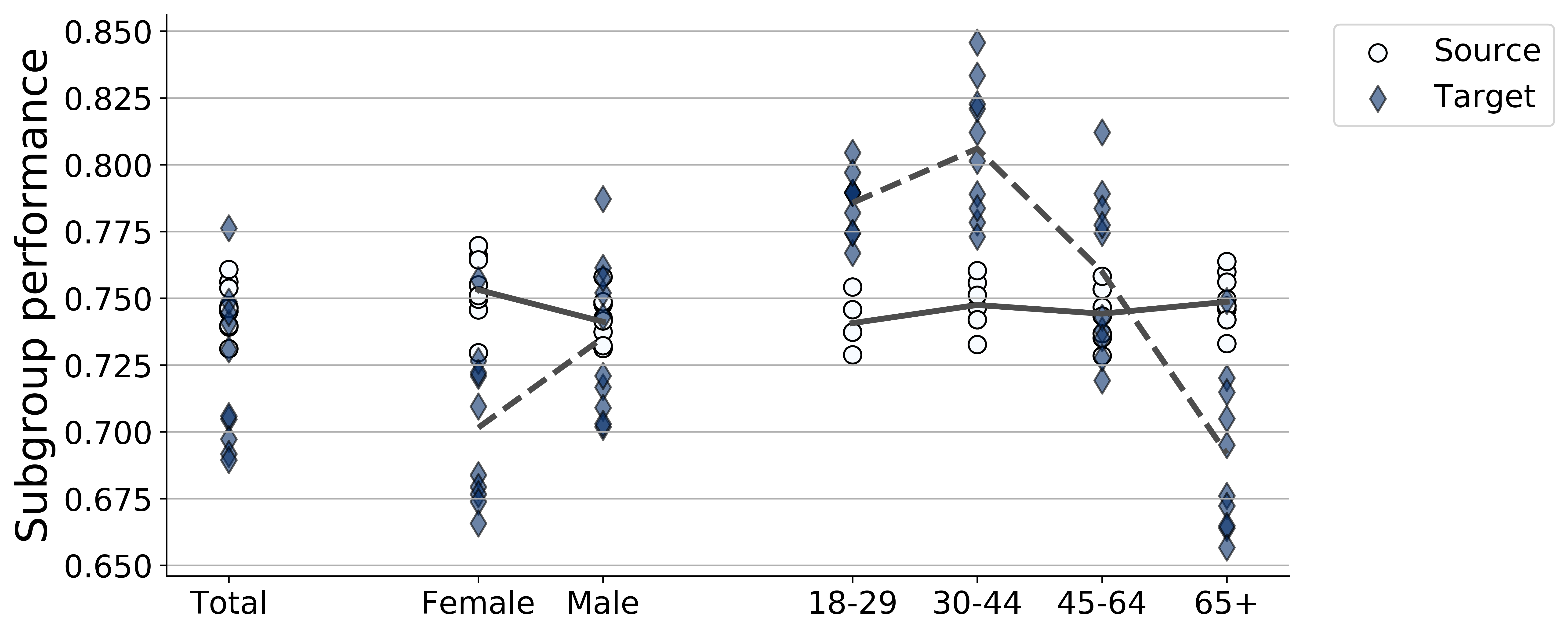}
\caption{EHR model accuracy across subgroups after post-processing of the predictions for equalized odds based on age.}
\label{fig:ehr_mitigated_eo}     
\end{figure*}

Based on this analysis, performing fairness mitigation in the presence of a compound shift would fail to provide a fair result on the target. Considering the applications above, a reasonable option would be to perform fairness mitigation for the EHR task based on age. We perform post-processing of the predictions to enforce either demographic parity or equalized odds across age subgroup using the method in \citet{alabdulmohsin2021}. This results in a decrease in demographic parity to value $10^{-5}$ in the source, but to (slightly) increase to value $0.078 \pm 0.025$ in the target data. Similarly, equalized odds decrease to value $10^{-5}$ and the gap in model accuracy between groups is reduced from 7\% to 0.8\% (Fig.~\ref{fig:ehr_mitigated_eo}) in the source, but increases from $0.05 \pm 0.008$ to $0.16 \pm 0.04$ on the target data, and the gap between age groups increases to 11.4\%. We therefore observe that mitigating for either demographic parity or equalized odds on the source leads to no improvement, or even worsening of the fairness properties on the target, as predicted by our analysis.

\noindent \textbf{Robustness mitigation: }Relations between robustness guarantees and individual and group fairness have been studied in \citep{Yurochkin2019-db,Nanda2021-tw,Xu2021-ze,Yeom2020-th} and \citep{Veitch2021-rq} respectively. \citep{Adragna2020-fs,Creager2021-ia,Zhang2021-bb} have investigated the relationship between fairness and distribution shift experimentally by studying how robustness mitigation strategies affect fairness metrics. These works do not provide provable robustly fair models.

\noindent \textbf{Robustly fair mitigation.} Work explicitly investigating fairness properties under distribution shift can mostly be divided into methods that provide a priori \textit{guarantees} or \textit{bounds} \citep[e.g. ][]{Schumann2019-nu,Singh2021-og,Subbaswamy2020-mi} and methods that \textit{adapt} to the new distribution \citep[e.g. ][]{coston2019fair,oneto2019learning,Slack2020-zz,zhao2020fair,kallus2018residual,du2021fair,taskesen2020distributionally}. 

Recent work by \citet{Singh2021-og} suggests a straightforward mitigation with guarantees, based on selecting features that satisfy the fairness transfer criteria outlined in Section~\ref{sec:background}.
However, this approach is limited under compound shifts with complex data.
Most notably, it cannot be used in cases where there is a label shift ($S \rightarrow Y$).
However, even when there is no label shift, under compound shifts the criteria for feature selection can yield a trivial predictor where no observed variables can be used.
This is the case in our EHR example under the joint causal graph in Fig.~\ref{fig:ehr_shift}(a).
We discuss the applicability of \citep{Singh2021-og} across different causal scenarios in the Supplement. In addition, feature selection approaches are more suited for semantic features and, while it is technically feasible, would not lead to insightful results in non-semantic features such as pixels.
In computer vision, masking has been used previously as an approximation to feature selection to constrain the learning towards specific areas of the image \citep{Ross2017-br,Simpson2019-zh}. 
While this technique has shown some promise with chest x-rays \citep{Viviano2019-ij}, it is impractical due to the additional requirement of a mask for all training images. In addition, it increases the ratio of signal of interest compared to undesirable signal, but does not exclude such signals and comes with no guarantee.

Another line of work considers adapting transfer learning and meta-learning algorithms to consider fairness constraints and therefore ensure generalization to new tasks also in terms of fairness \citep{coston2019fair,oneto2019learning,Slack2020-zz,zhao2020fair}. Specifically, \citet{coston2019fair} modify the standard approach to transfer learning under covariate shift by adding a constraint to the weight learning objective that enforces closeness between all pairs of protected groups. \citet{Schumann2019-nu} provide bounds on fairness transfer by adding adversarial heads on both the in-source attribute $A$ and the environment $S$. When we cast this work into our causal framework, we notice that it would provide a robustly fair model under an exclusive covariate shift in an anti-causal learning task.
\citet{Slack2020-zz} propose meta-learning with a regularizer that imposes demographic parity or equalized odds to improve the transfer of fairness. The method requires a small set of labelled target data (K-shot learning) and multiple tasks, and comes with no guarantees. Work on fine-tuning and transfer learning \citep{oneto2019learning, zhao2020fair} leverages task similarity to learn fair representations that provably generalize to unseen tasks, similarly requiring a small set of labelled target data. Importantly, these methods generate one tuned model per environment, which is not suitable for all real-world deployments of ML. It is also unclear whether these techniques could handle more complex tasks, e.g. predictions from sparse multivariate timeseries as in the case of EHR.

\noindent \textbf{Shift detection. } Many methods exist to detect covariate shift \citep[e.g.][]{Markou2003-yh,Chandola2009-ss} or label shift \citep{Lipton2018-ho,Garg2020-ul}. They however often rely on the assumption that other variables are not affected by the shift. For instance, \citet{Lipton2018-ho} relies on the assumption that $X \ind S \,|\, Y$. On the other hand, \citet{Slack2020-zz} introduce a technique to identify mean shifts that could lead to unfairness. While the method is an interesting step towards `fairness transfer warnings', this strategy does not allow to guide the selection of mitigation. Recently, \citet{Singh2022-na} investigate differences in source and target distributions using squared maximum mean discrepancy \citep{Gretton2012-ou}, but do not take the causal structure into account. To differentiate direct from indirect effects of $S$ on variables $U$, they perform causal inference and visually inspect the obtained graph. This method can only be applied to low-dimensionality datasets.



\section{Discussion}
In this work, we applied statistical tests within a joint causal framework \citep{Mooij2020-sv} to better understand failures of fairness transfer across distribution shift. Our approach tests for direct effects of the environment on every random variable in a task. We use this technique to characterize distribution shifts that result in two real world failures of fairness transfer in the healthcare domain. We show that this knowledge can then guide the choice of an appropriate mitigation strategy, when applicable.

Our work highlights real-world challenges that prevent the development of robustly fair models. Given that each task leads to different complexities in terms of causal graph, expected shifts, and fairness requirements, these findings support looking beyond purely algorithmic solutions. Considering the whole ML pipeline, we can re-interpret the remedies discussed in \citet{Chen2021-oc} for fairness transfer:
\begin{enumerate}[noitemsep,nolistsep]
    \item \textbf{Problem selection.} Consider focusing on tasks with lower anticipated harms or clinical/policy safeguards against ML unfairness.
    \item \textbf{Data collection.} Consider early data collection from anticipated deployment environments in which distribution shifts are suspected. This would allow to perform (conditional) independence tests to guide further mitigation.
    \item \textbf{Outcome definition.} Consider intermediate outcomes for which unfairness under distribution shift might be less consequential (e.g. image segmentation compared to diagnosis).
    \item \textbf{Algorithm development.} Based on (2), some strategies can be used if their assumptions are satisfied. In addition, in the present work we used a causal framework. However, techniques that are not easily cast into this framework exist. For instance, manual feature engineering has been used to improve the robustness of EHR models to distribution shift \citep{Nestor2019-dp}. While this work has not evaluated the fairness of the model in the source and target distributions, we believe that feature engineering might be a promising avenue to enforce specific inductive biases during learning while reducing the risks of distribution shifts. We however note that our work does not assess the impact of model architecture or different training strategies, which might lead to compounded bias and/or poor generalization.
    \item \textbf{Post-deployment.} Further safeguards may be envisaged to safely deploy ML systems, e.g. with a prospective observational (non-interventional) integration \citep{Finlayson2021-zo}. Transfer learning or retraining  with site-specific updates might then be considered as necessary prior to interventional deployment. Finally, those deployments may be accompanied with robust post-market surveillance to continuously monitor fairness.
\end{enumerate}%


\noindent \textbf{Limitations and future work.} The power of our diagnosis method depends on the number and quality of the samples. Therefore, it should be perceived as displaying potentially problematic aspects of the data, rather than assessing invariant features. This is a common limitation in shift detection \citep{Lipton2018-ho} and mitigation \citep{Singh2021-og}. Furthermore, the data at hand usually represents a set of proxies for the underlying variables. Therefore, even if a shift seems to satisfy some invariance assumptions, there are no sharp guarantees \citep{Finkelstein2021-eo}. This is especially true for demographic factors: features such as ethnicity, age or sex only approximate sensitive attributes. We are also limited by which factors are observed across all environments and cannot assess changes in unobserved variables (e.g. social determinants of health). 

While our work covers different types of shift, future work should investigate further sources of shift and biases, such as sampling biases or missingness \citep{castro2020causality,Zhang2021-bb}. In addition, extending the testing strategy to compare multiple (discrete or continuous) versions of the environment $S$ would be valuable.

By specifying causal graphs for each problem, we make assumptions about the relationships between variables. These assumptions might not represent the true underlying data generation process. 
In addition, the effects of $A$ could potentially be divided into `fair' and `unfair' effects, as proposed in \citep{Chiappa2019-kk,Mhasawade2020-ex}. For real-world applications, further assumptions were made, e.g. considering the silver standard label in dermatology as a good approximation for the gold standard label (hence an anti-causal task) \citep{castro2020causality}. The value of these graphs is purely illustrative and should not be considered for medical applications without validation.


In terms of metrics, we have focused mostly on demographic parity and equalized odds, due to their recent causal grounding \citep{Veitch2021-rq}. Different metrics might, however, be affected differently, and it is possible that some metrics are more or less `sensitive' to distribution shift. Similarly, different fairness mitigation strategies might be affected differently by distribution shift, although none provides guarantees under distribution shift. We also note that `fairness' in the healthcare domain is an active discussion, and that equalized odds or demographic parity metrics do not consider factors such as social determinants of health or health equity. As discussed in \citep{Pfohl2021-xn}, considering human elicited metrics might be an avenue forward. Similarly, `fairness' might change depending on the context \citep{Sambasivan2021-ba} and different metrics might be relevant in different environments.

\begin{ack}
The authors would like to acknowledge and thank Lucas Dixon, Noah Broestl, Sara Mahdavi, Nenad Tomasev, Cameron Chen, Stephen Pfohl, Matt Kusner, Victor Veitch, Jon Deaton, Shannon Sequeira, Abhijit Guha Roy, Jan Freyberg, Aaron Loh, Martin Seneviratne, Patricia MacWilliams, Yun Liu, Christopher Semturs, Dale Webster, Greg Corrado and Marian Croak for their contributions to this effort. This work was funded by Google.
\end{ack}

\bibliographystyle{plainnat}
\bibliography{refs}



\begin{enumerate}

\item For all authors...
\begin{enumerate}
  \item Do the main claims made in the abstract and introduction accurately reflect the paper's contributions and scope?
    \answerYes{We clearly describe that our main contribution is to provide a method that assesses the structure of a distribution shift.}
  \item Did you describe the limitations of your work?
    \answerYes{Please see the separate subsection in the Discussion.}
  \item Did you discuss any potential negative societal impacts of your work?
    \answerYes{In the limitation section, we discuss how algorithmic fairness is a set of mathematical formulations and does not ensure fairness in outcomes or health equity.}
  \item Have you read the ethics review guidelines and ensured that your paper conforms to them?
    \answerYes{}
\end{enumerate}

\item If you are including theoretical results...
\begin{enumerate}
  \item Did you state the full set of assumptions of all theoretical results?
    \answerYes{Assumptions are most relevant for our statistical tests and the simplified causal graphs we propose. We clearly mention these in the text, as well as in the discussion.}
        \item Did you include complete proofs of all theoretical results?
    \answerNA{}
\end{enumerate}

\item If you ran experiments...
\begin{enumerate}
  \item Did you include the code, data, and instructions needed to reproduce the main experimental results (either in the supplemental material or as a URL)?
    \answerYes{The main contribution of this work is the statistical tests. Algorithm \ref{alg} and the Supplement provide detailed instructions to re-implement the approach, including practical tips.}
  \item Did you specify all the training details (e.g., data splits, hyperparameters, how they were chosen)?
    \answerYes{}
        \item Did you report error bars (e.g., with respect to the random seed after running experiments multiple times)?
    \answerYes{We report our results on 10 seeds of each model.}
        \item Did you include the total amount of compute and the type of resources used (e.g., type of GPUs, internal cluster, or cloud provider)?
    \answerYes{Computational resources are discussed in the Supplement.}
\end{enumerate}

\item If you are using existing assets (e.g., code, data, models) or curating/releasing new assets...
\begin{enumerate}
  \item If your work uses existing assets, did you cite the creators?
    \answerYes{We cite the creators of both healthcare datasets and refer to the code we used or took inspiration from throughout the text and in the Supplement.}
  \item Did you mention the license of the assets?
    \answerYes{Where available.}
  \item Did you include any new assets either in the supplemental material or as a URL?
    \answerNo{}
  \item Did you discuss whether and how consent was obtained from people whose data you're using/curating?
    \answerNo{}
  \item Did you discuss whether the data you are using/curating contains personally identifiable information or offensive content?
    \answerYes{See Supplement.}
\end{enumerate}

\item If you used crowdsourcing or conducted research with human subjects...
\begin{enumerate}
  \item Did you include the full text of instructions given to participants and screenshots, if applicable?
    \answerNA{}
  \item Did you describe any potential participant risks, with links to Institutional Review Board (IRB) approvals, if applicable?
    \answerYes{We discuss the ethical reviews for the dermatology datasets in the Supplement.}
  \item Did you include the estimated hourly wage paid to participants and the total amount spent on participant compensation?
    \answerNA{}
\end{enumerate}

\end{enumerate}


\newpage
\appendix
\section*{Appendix}

\section{Methods}
In this section, we provide further details on the methods, including the mathematical definitions of our metrics and an in-depth description of our testing procedure.

\subsection{Fairness metrics}
\label{app:fairness_metrics}

Our work focuses on statistical group definitions of fairness \citep{Barocas2019-kj}. In particular, we frame the discussion around demographic parity in which the model's output is statistically independent of the sensitive attribute, i.e. $f(X')\,\ind\, A$, and equalized odds in which the independence holds conditionally on the outcome, i.e. $f(X')\,\ind\,A\, |\, Y$. In the experiments, these metrics are computed for binary tasks. We also compute the gap in subgroup accuracy. For all fairness metrics, the lower, the better.

We define demographic parity for a predictor $f$ as \citep{dwork2012fairness,zafar2017fairness,mehrabi2019survey,alabdulmohsin2021}:
\begin{equation}
 \max_{a\in\mathcal{A}}\,\mathbb{E}_{\bx}[f(\bx)\,|\,A=a] \,-\,\min_{a\in\mathcal{A}}\,\mathbb{E}_{\bx}[f(\bx)\,|\,A=a]
\end{equation}

Equalized odds is computed in a similar fashion by conditioning on the positive and negative classes, taking the average of the discrepancies across classes.

Subgroup performance is computed as the accuracy (top-1 or top-3 in the case of dermatology, accuracy for EHR) within each subgroup. When the sensitive attribute is non-binary, we compute the maximum difference between any pair of subgroups.

\subsection{Causal framework}
\label{app:methods_causal_framing}
A \emph{Bayesian network} \citep{cowell2001probabilistic,kollerl2009probabilistic,pearl1988probabilistic,pearl2000causality} is a \emph{directed acyclic graph} (DAG) ${\mathcal G}$ whose nodes $X^1,\ldots, X^D$ represent random variables and links express statistical dependencies among them. Each node $X^d$ is associated with a \emph{conditional probability distribution} (CPD)~$p(X^d \,|\, \pa(X^d))$, where $\pa(X^d)$ denote the \emph{parents} of $X^d$, namely the nodes with a link into $X^d$. The joint distribution of all nodes is given by the product of all CPDs, i.e. $p(X^1, \dots, X^D \,|\,{\mathcal G}) = \prod_{d=1}^D p(X^d \,|\, \pa(X^d))$. This function is assumed to be invariant across distributions. A \emph{path} from $X_i$ to $X_j$ is a sequence of linked nodes starting at $X_i$ and ending at $X_j$. A path is called \emph{directed} if the links point from preceding towards following nodes in the sequence. A node $X_i$ is an \emph{ancestor} of a node $X_j$ if there exists a directed path from $X_i$ to $X_j$. In this case, $X_j$ is a \emph{descendant} of $X_i$. We say that a set of nodes \{$X^i,\ldots,X^j$\} \emph{d-separates} two random variables $U$ and $W$ if $U \ind W \,|\, {\{X^i,\ldots,X^j\}}$. A causal Bayesian network is a Bayesian network in which a link expresses causal influence rather than statistical dependence. In causal Bayesian networks, directed paths are called \emph{causal paths}.

\subsection{Assessing the causal structure of shifts}
\label{app:methods_shift}
\subsubsection{General Strategy}
To assess the causal structure of a shift, we examine whether there is a direct effect of the shift $S$ on a focal variable $U$.
This requires controlling for all other pathways by which $U$ may depend on $S$.
To do this, we select a set of variables $\mathbf V$ such that \{$\mathbf V$\} blocks all indirect paths from $S$ to $U$.
We then test the following equality in conditional distributions: $P(U \mid \mathbf{V}, S=0) = P(U \mid \mathbf{V}, S=1)$ almost everywhere.

To test the equality of conditional distributions, we reduce the problem to testing whether the means of the two marginal distributions of $U$ are equal when the distributions of the background variables $\mathbf{V}$ are adjusted to follow the same distribution.
Using observed data from environments $S=0$ and $S=1$, we can perform such an apples-to-apples comparison using importance weighting.
In particular, for any distribution $\pi(\mathbf{V})$,
\begin{align}
P(U | \mathbf{V}, S=0) &= P(U | \mathbf{V}, S=1) \quad\text{for almost all }\mathbf{V} \quad \implies\\
\int E[U \mid \mathbf{V} \mid S=0] \pi(\mathbf{V}) d\mathbf{V} &= \int E[U \mid \mathbf{V} \mid S=1] \pi(\mathbf{V}) d\mathbf{V} \quad \implies\\
E\left[\frac{\pi(\mathbf{V})}{P(\mathbf{V} \mid S=0)} E[U \mid \mathbf{V} \mid S=0] \mid S=0\right] &= E\left[\frac{\pi(\mathbf{V})}{P(\mathbf{V} \mid S=1)} E[U \mid \mathbf{V} \mid S=1] \mid S=1\right] \implies \\ 
E\left[\frac{\pi(\mathbf{V})}{P(\mathbf{V} \mid S=0)} U \mid S=0\right] &= E\left[\frac{\pi(\mathbf{V})}{P(\mathbf{V} \mid S=1)} U \mid S=1\right]
\end{align}

Define the weights $w_0(\mathbf{V}) := \frac{\pi(\mathbf{V})}{P(\mathbf{V} \mid S=0)}$, and $w_1(\mathbf{V}) := \frac{\pi(\mathbf{V})}{P(\mathbf{V} \mid S=1)}$.
This result shows that we can test the equality of the conditional distributions by testing the implication that the mean of weighted outcomes is the same.

This leaves a degree of freedom for choosing the test distribution $\pi(\mathbf{V})$ on which the conditional distributions will be compared.
This distribution then determines the weighting scheme that will be used.
We define $\pi(\mathbf{V})$ to be a constant, i.e., uniform.
Then the weights are
$$w_0(\mathbf{V}) \propto P(\mathbf{V} \mid S=0)^{-1} \quad\text{and}\quad w_1(\mathbf{V}) \propto P(\mathbf{V} \mid S=1)^{-1}.$$
This weighting scheme is known as inverse probability weighting or IPW \citep{Rosenbaum1983-zd,Imai2004-uh}, and is a popular choice. We however note that other weighting schemes could be considered, e.g. overlap weights \citep{Li2019-nv} or permutation weighting \citep{Arbour2021-fm}.



In practice, the likelihoods $P(\mathbf{V} \mid S=s)$ used to define the weights need to be learned from data.
Usefully, because of how the weights are normalized, the likelihoods can be replaced with classification scores that estimate $P(S = s \mid \mathbf{V})$, leading to estimated weights $\hat w_0$ and $\hat w_1$.
The choice of the classifier can be informed by the expected relationships between \{$\mathbf V$\} and $S$: for instance, logistic regression can be considered in the case of simple relationships, while more expressive classifiers such as gradient boosted trees can be considered in the case of non-linear relationships \citep{Arbour2021-fm}. As discussed in \citet{Arbour2021-fm}, all classifiers are tuned (C for logistic regression, number of trees, tree depth and learning rate for boosted trees) in a cross-validation setup.

We then use a standard t-test to test the null hypothesis $$H_0: E[\hat w_0 U \mid S=0] - E[\hat w_1 U \mid S=1] =0.$$

To decrease the variance in the obtained t-statistics, we bootstrap the weighting and testing procedure. The final p-value associated with $H_0: P(U \mid \mathbf{V}, S=0) = P(U \mid \mathbf{V}, S=1)$ is the p-value of a two-sided $\mathcal{Z}$-test on the t-statistics obtained from each bootstrap.

\noindent \textbf{Observations and tips }From our observations, we note that it is best to allocate the largest available sample to estimate the likelihood ratio, compared to performing the final statistical test. We also observed that gradient boosted trees can lead to extreme weights in some cases. This results in an increased variance and non-significant results. We therefore clipped the weights to a maximum of 10 (arbitrary choice) to mitigate this effect.

\subsubsection{Testing with High-Dimensional $U$}

If $U$ has low dimensionality, \citet{Rabanser2019-gu} show that multiple one-dimensional tests can be used with correction for multiple comparisons (here Bonferroni).
On the other hand, if $U$ has higher dimensionality (e.g. an image), summaries of $U$ can be constructed and the test can be performed on these lower-dimensional summaries.
The validity of this approach follows from the fact that if the conditional distributions of $U$ are the same, then so will the conditional distributions of any summary $f(U)$.
The trade-off is that the test loses all power to detect distributional differences that are compressed out by the summary $f(U)$, or to highlight specific dimensions in $U$ that are more or less affected by $S$.
To define a summary that is relevant to the problem, \citet{Lipton2018-ho} suggest defining $f(U)$ to be the output of a model the predicts some variable of interest (say the outcome $Y$) using $U$.
As reported in \citet{Rabanser2019-gu}, other summarizing techniques could be considered.

\subsubsection{Sanity checks with engineered shifts}
\label{app:methods_checking_tests}
We assess the specificity and sensitivity of our testing procedure (see Algorithm 1) using engineered shifts on the dermatology data. To estimate Type I error, we compare the source data to itself, which should lead to all variables being independent of the environment. More specifically, we select $10,000$ random samples from the dataset to compute the weights, and another $1,000$ random samples to apply the weights on and perform the weighted test. These sets are boostrapped 100 times (separately for the source and target sets) and we assess the proportion of false positives that is obtained for each variable $U$. We expect that we will obtain $~5\%$ of false positives, as defined by our hypothesis testing threshold. As we do not expect any relationship between  \{$\mathbf V$\} and $S$, we use a logistic regression to estimate $P(S = s \mid \mathbf{V})$.

\begin{figure*}[!ht]
\centering
\begin{subfigure}[t]{0.03\textwidth}
(a)
\end{subfigure}
\begin{subfigure}[t]{0.45\textwidth}
\includegraphics[width=\linewidth,valign=t]{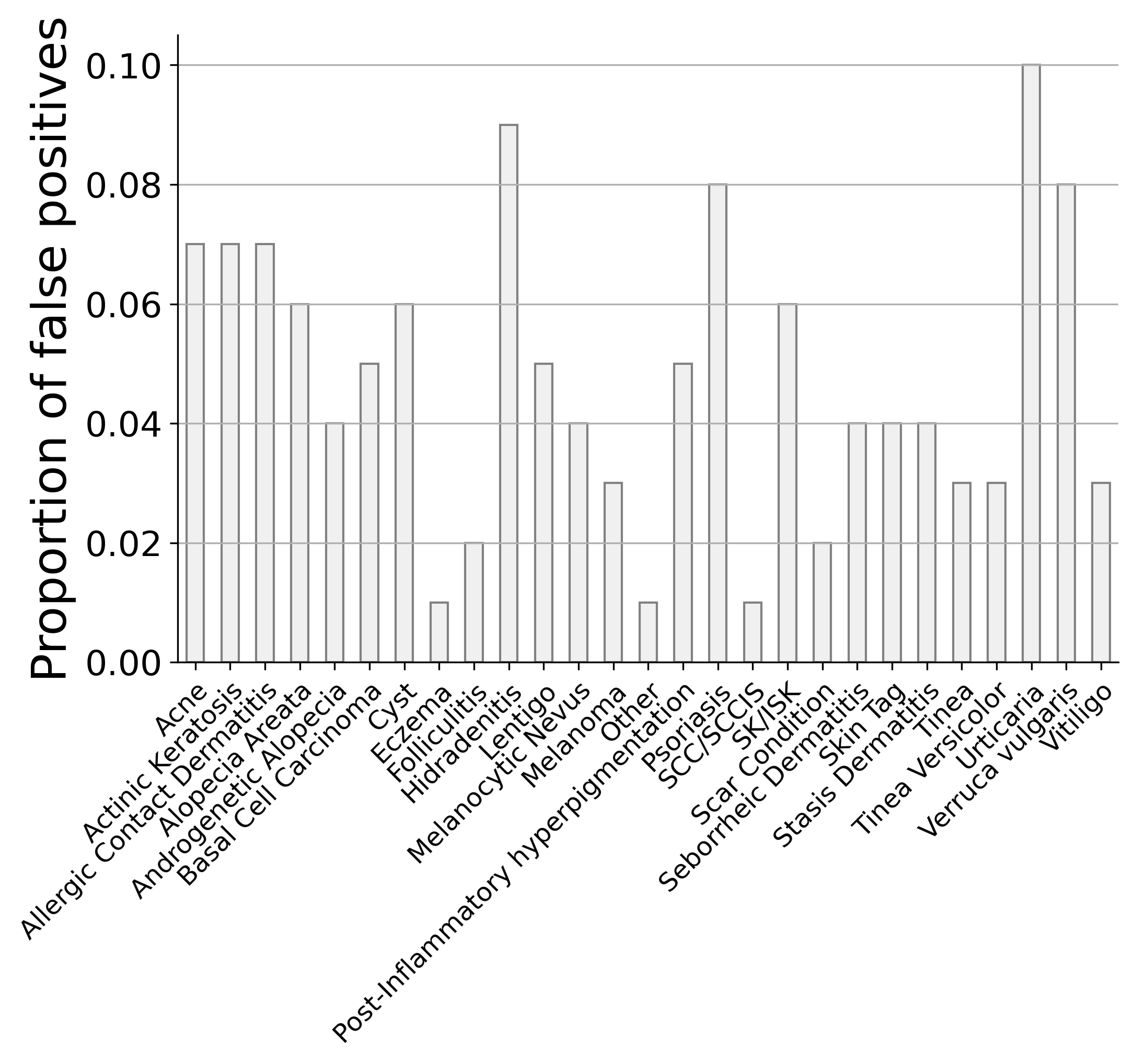} 
\end{subfigure}
\begin{subfigure}[t]{0.03\textwidth}
(b)
\end{subfigure}
\begin{subfigure}[t]{0.45\textwidth}
\includegraphics[width=\linewidth,valign=t]{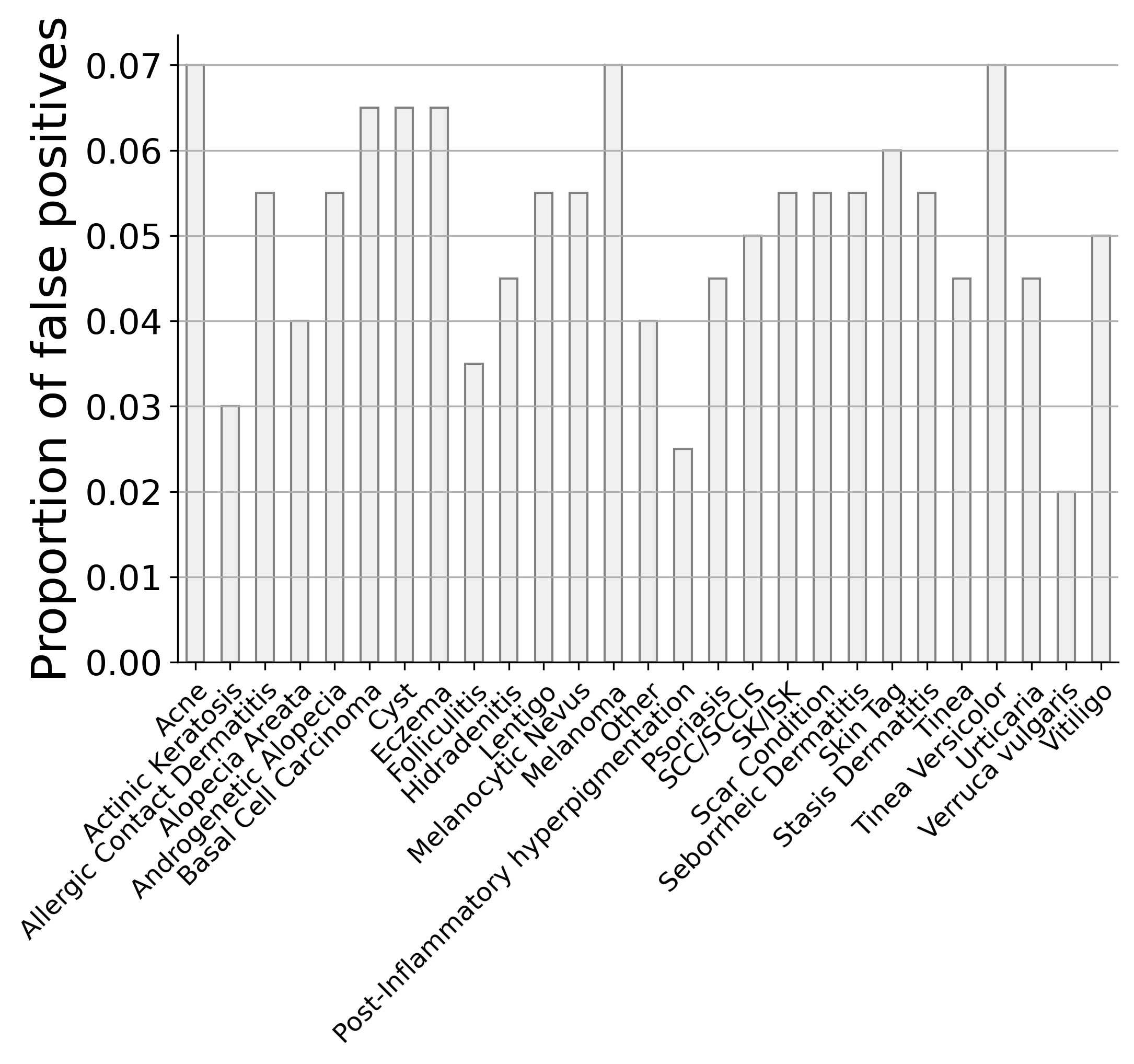}
\end{subfigure}\\
\begin{subfigure}[t]{0.03\textwidth}
(c)
\end{subfigure}
\begin{subfigure}[t]{0.6\textwidth}
\includegraphics[width=\linewidth,valign=t]{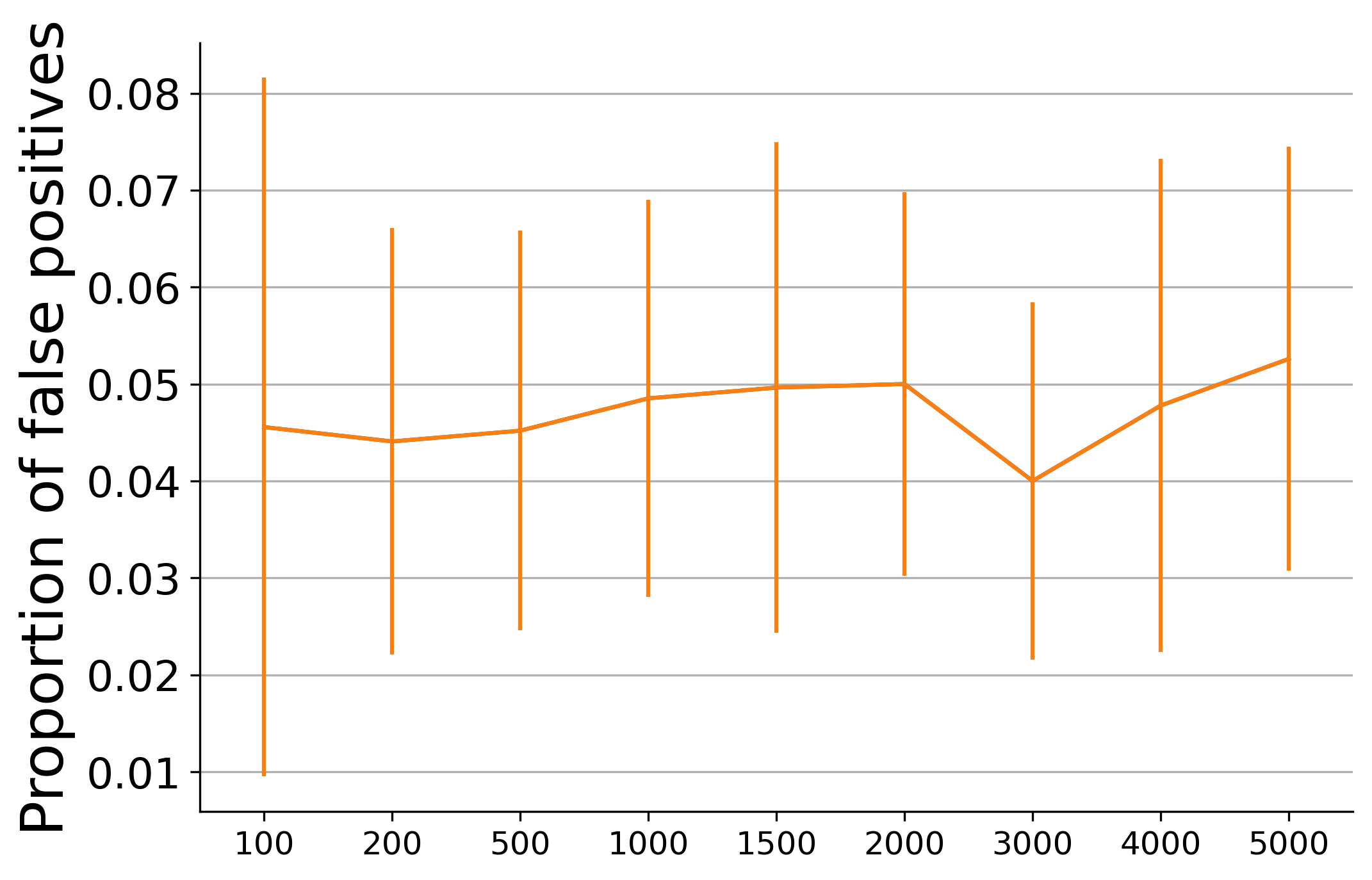}
\end{subfigure}
\caption{Proportions of false positives when comparing the source data to itself. (a) $P(Y \mid A, S=0)$ vs $P(Y \mid A, S=1)$. (b) $P(X \mid Y, A, S=0)$ vs $P(X \mid Y, A, S=1)$. (c) $P(Y \mid A, S=0)$ vs $P(Y \mid A, S=1)$ when varying the number of samples to estimate $P(S \mid \{\mathbf{V}\})$. In this case, the results are averaged across conditions (with standard deviation). }
\label{fig:methods_sanity_check}       
\end{figure*}

When comparing $P(Y \mid A, S=0)$ to $P(Y \mid A, S=1)$, we observe that the accuracy of the classifier mirrors the chance level of $50\%$ and that the proportion of false positives varies between $2$ and $8.5\%$ (Figure~\ref{fig:methods_sanity_check}(a)). Similarly, comparing $P(X \mid Y, A, S=0)$ to $P(X \mid Y, A, S=1)$ when using a model $f(X) \rightarrow Y$ to summarize $X$ leads to a proportion of false positives between 2 and $7\%$ (Figure~\ref{fig:methods_sanity_check}(b)). 

As the number of samples used to compute the weights is an important component of our testing approach, we repeat the process while varying the number of samples from 100 to 5,000. Figure~\ref{fig:methods_sanity_check}(c) displays that the variance on Type I error (here across conditions) decreases when the number of samples to fit the $P(S = s \mid \mathbf{V})$ classifier increases.

To assess sensitivity, we then engineer a shift on the data that discards samples based on their age and condition (here Acne). More specifically, we subsample data in order to decrease the bias for younger age in the Acne condition, leading to an increase in age for Acne samples of approximately 10 years. Given that Acne is one of the most prevalent conditions in the dataset, we expect a significant change on age, as well as on $Y$. On the other hand, the images are not modified and hence $P(X \mid Y, A, S=0)$ and $P(X \mid Y, A, S=1)$ should not display significant changes.

We observe a significant change in age between $S=0$ and $S=1$. There is no significant effect of $S$ on other demographic factors. We do observe a direct effect of $S$ on $Y$, with a strong effect on Acne as expected. On the other hand, no tests display significant differences when comparing $P(X \mid Y, A, S=0)$ with $P(X \mid Y, A, S=1)$. This latter result was obtained with both a logistic regression and gradient boosted trees (given the more complex relationship between age and Acne). Therefore, the testing approaches produces a causal graph that mirrors the graph used to engineer the shift, and our procedure is considered faithful.

Finally, we randomly shuffled the labels $y_k$ across the 27 conditions, for each case independently. At the population level, this results in a change in $P(Y)$. As expected, there was no effect of $S$ on $A$. For $Y$, we observed significant differences in $P(Y \mid A, S=0)$ and $P(Y \mid A, S=1)$ for 19 conditions out of 27 (Bonferroni corrected). Thanks to the correction, and given that the images were not modified, there are no significant differences between $P(X | Y, A, S=0)$ and $P(X | Y, A, S=1)$.

Overall, our tests are sensitive enough to detect the presence of changes on the affected dimensions, while not displaying an excessive amount of false positives.

\section{Causal framing of related works}
\label{app:rel_work}
In this section, we review how various transferable fairness approaches that have been proposed in the literature interact with different causal structures of domain shift depicted in Figures~\ref{fig:anticausal_graphs} and \ref{fig:causal_graphs}.
We assume the same sensitive attributes $A$ in the source and target environments, and consider different relationships between the environment and other variables. As in \citep{castro2020causality,Veitch2021-rq}, we split the analysis into causal or anti-causal prediction tasks \citep{scholkopf2012on}.
A prediction task is causal if the features $X$ are causes of the outcome $Y$, and anti-causal if the outcome $Y$ is a cause of the features $X$.
Many risk prediction problems are causal, where risk factors may cause an adverse outcome, while many diagnostic problems are anti-causal, where the underlying disease may cause symptoms \citep{castro2020causality}.

A takeaway from this review is that most methods are tailored to particular restricted shift structures, and their useful properties can break down under compound shifts.

\subsection{Anti-causal prediction tasks}\label{sect::anticausal_struct}
First, we consider anti-causal prediction tasks, in which the outcome $Y$ is a direct cause of the features $X$, and that the sensitive attribute $A$ is a direct cause of $X$ and $Y$ (the reality will likely be more complex and involve other variables in the paths (observed or not)). The assumption that $A$ is a cause of $X$ and $Y$ is a reasonable approximation in healthcare settings: e.g. on average, skin images from men are more hairy than those of women, breasts are visible in chest x-rays, and comorbidities are different across sexes and age ranges. Sensitive attributes can also be related to the label, as prevalence can vary across subgroups (e.g. baldness is more prevalent in older patients) and some conditions might present differently across attributes (e.g. a heart attack in men vs women).

In this context, we consider strategies for learning robustly fair models under exclusive demographic, exclusive covariate, exclusive label and compound distribution shift scenarios represented in the causal graphs of Fig.~\ref{fig:anticausal_graphs}:

\begin{figure}
\centering
\begin{subfigure}{\textwidth}
\hspace{0.07\columnwidth}(a)\hspace{0.23\columnwidth}(b)\hspace{0.23\columnwidth}(c)\hspace{0.23\columnwidth}(d)\\
\scalebox{1}{
\begin{tikzpicture}
\node (S) at (-1.5,0) {$S$};
\node (A) at (-1.5,-1) {$A$};
\node (Y) at (-1.5,-2) {$Y$};
\node (X) at (1,-2) {$X$};
\draw[line width=1pt,blue, \arr, opacity=0.6](S)--(A);
\draw[line width=1pt,black, \arr, opacity=0.7](A)--(X);
\draw[line width=1pt,black, \arr, opacity=0.7](Y)--(X);
\draw[line width=1pt,black, \arr, opacity=0.7](A)--(Y);
\node[] (I1) at (-0.45,-2.6) {$X \,\ind\, S \;|\; A$};
\node[] (I2) at (-0.30,-3) {$X \,\ind\, S \;|\; A,Y$};
\node[] (I3) at (-0.45,-3.4) {$Y \,\ind\, S \;|\; A$};
\node[draw,dotted,fit=(I1)(I2)(I3)] {};
\end{tikzpicture}}
\scalebox{1}{
\begin{tikzpicture}
\node (S) at (0,-0.5) {$S$};
\node (A) at (-1.5,-1) {$A$};
\node (Y) at (-1.5,-2) {$Y$};
\node (X) at (1,-2) {$X$};
\draw[line width=1pt,black, \arr, opacity=0.7](A)--(Y);
\draw[line width=1pt,black,\arr, opacity=0.7](Y)--(X);
\draw[line width=1pt, \arr, opacity=0.7](A)--(X);
\draw[line width=1pt,blue, \arr, opacity=0.6](S)--(X);
\node[] (I1) at (-0.4,-2.6) {$A \ind S$};
\node[] (I2) at (-0.35,-3) {$Y \,\ind\, S$};
\node[] (I3) at (-0.4,-3.65) {};
\node[draw,dotted,fit=(I1)(I2)] {};
\end{tikzpicture}}
\scalebox{1}{
\begin{tikzpicture}
\node (S) at (-2.5,-1.5) {$S$};
\node (A) at (-1.5,-1) {$A$};
\node (Y) at (-1.5,-2) {$Y$};
\node (X) at (1,-2) {$X$};
\draw[line width=1pt,black, \arr, opacity=0.7](A)--(Y);
\draw[line width=1pt,black, \arr, opacity=0.7](Y)--(X);
\draw[line width=1pt, \arr, opacity=0.7](A)--(X);
\draw[line width=1pt,blue, \arr, opacity=0.6](S)--(Y);
\node[] (I1) at (-0.6,-2.6) {$A \ind S$};
\node[] (I2) at (-0.3,-3) {$X \,\ind\, S \;|\; Y$};
\node[] (I3) at (-0.4,-3.65) {};
\node[draw,dotted,fit=(I1)(I2)] {};
\end{tikzpicture}}
\scalebox{1}{
\begin{tikzpicture}
\node (S) at (0,-0.5) {$S$};
\node (A) at (-1.5,-1) {$A$};
\node (Y) at (-1.5,-2) {$Y$};
\node (X) at (1,-2) {$X$};
\draw[line width=1pt,black, \arr, opacity=0.7](A)--(X);
\draw[line width=1pt,black, \arr, opacity=0.7](Y)--(X);
\draw[line width=1pt,black, \arr, opacity=0.7](A)--(Y);
\draw[line width=1pt,blue, \arr, opacity=0.6](S)--(A);
\draw[line width=1pt,blue, \arr, opacity=0.6](S)--(Y);
\draw[line width=1pt,blue, \arr, opacity=0.6](S)--(X);
\node[] (I1) at (-0.55,-2.6) {};
\node[] (I2) at (-0.55,-3) {};
\node[] (I3) at (-0.4,-3.65) {};
\end{tikzpicture}}
\end{subfigure}
    \caption{Causal graphs under different distribution shifts when considering an anti-causal prediction task. (a) Exclusive demographic shift where the demographics $A$ are directly affected by $S$. (b) Exclusive covariate shift where the $X$ is directly affected by $S$. (c) Exclusive label shift where $Y$ is directly affected by $S$. (d) Compound shift given by the combination of demographic, covariate, and label shifts. 
    }
    \label{fig:anticausal_graphs}
\end{figure}
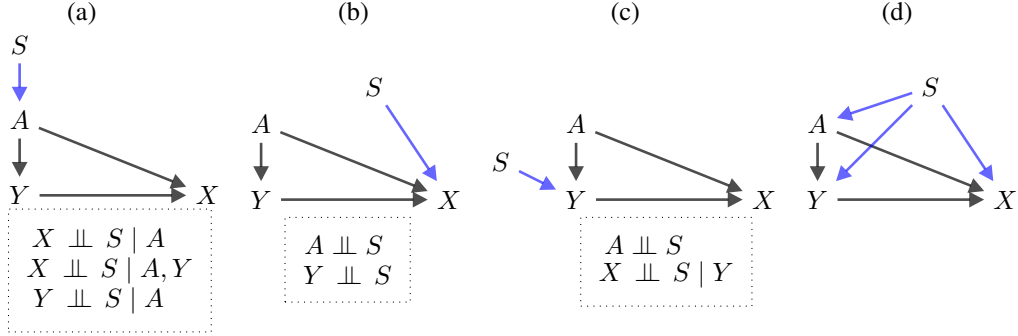

\textbf{a) Exclusive demographic shift.} In Fig.~\ref{fig:anticausal_graphs}(a), we consider an exclusive demographic shift through the direct path $S\rightarrow A$. In this case, the effects of $S$ on $Y$ and $X$ are intercepted by $A$. Therefore, training a fair model based on $X$ (either via imposing equalized odds \citep{Singh2021-og} or counterfactual invariance with respect to $A$ \citep{Veitch2021-rq}) leads to a robustly fair model. 

\textbf{b) Exclusive covariate shift.} In Fig.~\ref{fig:anticausal_graphs}(b), we consider an exclusive covariate shift through the direct path $S\rightarrow X$. In this case, there are multiple causal arrows into $X$ that need to be addressed: $A \rightarrow X$ for fairness, and term $S \rightarrow X$ for robustness to distribution shift. 
In the case where some labeled target domain data is available, \citet{Schumann2019-nu} implement separate independence regularizers to address each path, and demonstrate its effectiveness in settings that match this causal structure.
In the fully unlabeled domain adaptation case, the shift is harder to account for.
For example, the set of features that satisfy the selection criteria of \citet{Singh2021-og} would only include $A$, leading to a predictor that relies solely on demographics data (Sec.~\ref{app:methods_feat_sel}).

\textbf{c) Exclusive label shift.} \citet{Lipton2018-ho} propose a technique to detect an exclusive label shift, as well as mitigate it through reweighting, requiring unlabelled target data. This technique can be used in addition to a fairness mitigation technique to lead to a robustly fair model. 

\textbf{d) Compound shift.} Such a shift leads to multiple factors being affected by $S$, and current techniques would either be ineffective or degenerate to the trivial predictor. While Fig.~\ref{fig:anticausal_graphs}(d) represents the worst case scenario, even simpler combinations of shifts would lead to similar results in practice. For instance, adding a correlation between $S$ and $A$ to Fig.~\ref{fig:anticausal_graphs}(b) leads to a compound shift. In this specific case, mitigation might be possible by taking all intersections of $A$ and $S$ when using \citep{Schumann2019-nu}. In practice, if the attribute or the shift have multiple discrete values, this will quickly become intractable, especially if distribution matching needs to be applied within each mini-batch. When considering feature selection, \citep{Singh2021-og} would return an empty set for both equalized odds or demographic parity, leading to a trivial predictor. (Partial) mitigation might be obtained through the use of an adaptation technique as proposed in \citep{Slack2020-zz,oneto2019learning}.

\subsection{Causal prediction tasks}\label{sect::causal_struct}
We repeat the analysis above for the simplest causal prediction task in which $X$ is a direct cause of $Y$, and assuming that $A$ is a direct cause of $X$ and $Y$. We then consider the following distribution shift scenarios represented in Fig.~\ref{fig:causal_graphs}:

\begin{figure}
\centering
\begin{subfigure}{\textwidth}
\hspace{0.08\columnwidth}(a)\hspace{0.23\columnwidth}(b)\hspace{0.23\columnwidth}(c)\hspace{0.23\columnwidth}(d)\\
\scalebox{1}{
\begin{tikzpicture}
\node (S) at (-1.5,0) {$S$};
\node (A) at (-1.5,-1) {$A$};
\node (X) at (-1.5,-2) {$X$};
\node (Y) at (1,-2) {$Y$};
\draw[line width=1pt,blue, \arr, opacity=0.6](S)--(A);
\draw[line width=1pt,black, \arr, opacity=0.7](A)--(X);
\draw[line width=1pt,black, \arr, opacity=0.7](X)--(Y);
\draw[line width=1pt,black, \arr, opacity=0.7](A)--(Y);
\node[] (I1) at (-0.55,-2.6) {$X \,\ind\, S \;|\;A $};
\node[] (I2) at (-0.55,-3) {$Y \,\ind\, S \;|\; A$};
\node[] (I3) at (-0.4,-3.4) {$Y \,\ind\, S \;|\; A,X$};
\node[draw,dotted,fit=(I1) (I2) (I3)] {};
\end{tikzpicture}}
\scalebox{1}{
\begin{tikzpicture}
\node (S) at (-2.5,-1.5) {$S$};
\node (A) at (-1.5,-1) {$A$};
\node (X) at (-1.5,-2) {$X$};
\node (Y) at (1,-2) {$Y$};
\draw[line width=1pt,black, \arr, opacity=0.7](A)--(X);
\draw[line width=1pt,black, \arr, opacity=0.7](X)--(Y);
\draw[line width=1pt,black, \arr, opacity=0.7](A)--(Y);
\draw[line width=1pt,blue, \arr, opacity=0.6](S)--(X);
\node[] (I1) at (-0.4,-2.6) {$A \,\ind\, S $};
\node[] (I2) at (-0.2,-3) {$Y \,\ind\, S \;|\; X$};
\node[] (I3) at (-0.4,-3.65) {};
\node[draw,dotted,fit=(I1)(I2)] {};
\end{tikzpicture}}
\scalebox{1}{
\begin{tikzpicture}
\node (S) at (0,-0.5) {$S$};
\node (A) at (-1.5,-1) {$A$};
\node (X) at (-1.5,-2) {$X$};
\node (Y) at (1,-2) {$Y$};
\draw[line width=1pt,black, \arr, opacity=0.7](A)--(X);
\draw[line width=1pt,black, \arr, opacity=0.7](X)--(Y);
\draw[line width=1pt,black, \arr, opacity=0.7](A)--(Y);
\draw[line width=1pt,blue, \arr, opacity=0.6](S)--(Y);
\node[] (I1) at (-0.35,-2.6) {$A \,\ind\, S $};
\node[] (I2) at (-0.35,-3) {$X \,\ind\, S $};
\node[] (I3) at (-0.4,-3.65) {};
\node[draw,dotted,fit=(I1)(I2)] {};
\end{tikzpicture}}
\scalebox{1}{
\begin{tikzpicture}
\node (S) at (0,-0.5) {$S$};
\node (A) at (-1.5,-1) {$A$};
\node (X) at (-1.5,-2) {$X$};
\node (Y) at (1,-2) {$Y$};
\draw[line width=1pt,black, \arr, opacity=0.7](A)--(X);
\draw[line width=1pt,black, \arr, opacity=0.7](X)--(Y);
\draw[line width=1pt,black, \arr, opacity=0.7](A)--(Y);
\draw[line width=1pt,blue, \arr, opacity=0.6](S)--(A);
\draw[line width=1pt,blue, \arr, opacity=0.6](S)--(Y);
\draw[line width=1pt,blue, \arr, opacity=0.6](S)--(X);
\node[] (I1) at (-0.55,-2.6) {};
\node[] (I2) at (-0.55,-3) {};
\node[] (I3) at (-0.4,-3.65) {};
\end{tikzpicture}}
\end{subfigure}
\caption{Causal graphs under different distribution shifts when considering a causal prediction task. 
(a) Exclusive demographic shift, (b) Exclusive covariate Shift, (c) Exclusive label shift, and (d) Compound shift. 
}
\label{fig:causal_graphs}
\end{figure}
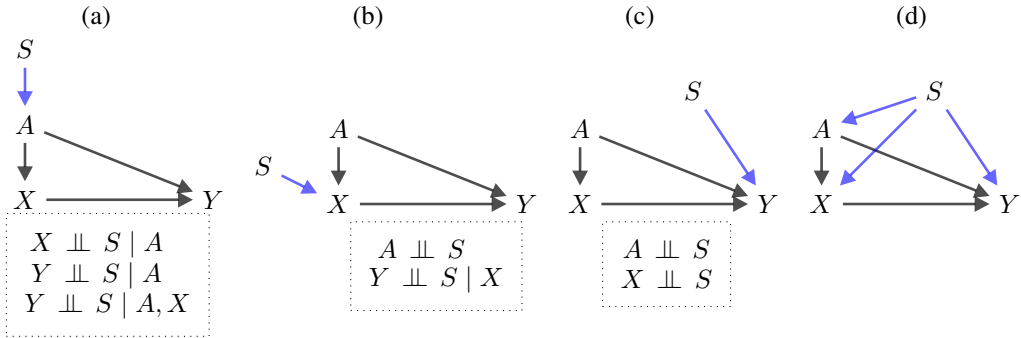

\textbf{a) Exclusive demographic shift.} As in the anti-causal case, the relationship between $X$ and $Y$ is not affected by $S$ if we impose invariance of $Y$ on $A$. To this end, training a fair model based on the set \{$X,A$\} would lead to a robustly fair model \citep{Singh2021-og}.

\textbf{b) Exclusive covariate shift.} In this setting, note that $(Y, A) \ind S \,|\, X$. Thus, fairness and robustness to distribution shift guarantees can be obtained independently.
This setting corresponds to the classic covariate shift scenario treated in much of the domain adaptation literature, where potential solutions are to match the distributions of $X$ across environments by strategies such as reweighting \citep[e.g.][]{shimodaira2000improving} or invariant representation learning \citep[e.g.][]{Ben-David2010-tv}. The advantage of such a techniques is that they only require unlabelled target data.

\textbf{c) Exclusive label shift.} While solutions are investigated for anti-causal prediction tasks, label shifts on causal prediction tasks are understudied and the absence of a direct path $S \rightarrow Y$ is often an assumption for mitigation techniques \citep{Singh2021-og,Veitch2021-rq}. To the best of our knowledge, there is no method proposing formal fairness guarantees under exclusive label shift in causal prediction tasks.

\textbf{d) Compound shift.} As for the anticausal case, a compound shift would lead to insufficient or trivial predictors.

\subsection{Feature selection mitigation strategies}
\label{app:methods_feat_sel}
We summarize the results for feature selection, and in particular the method proposed in \citep{Singh2021-og} in Table~\ref{tab:limitations_featsel}. We observe that exclusive covariate and label shifts cannot be mitigated without excluding the signal, leading to an empty set $\{\mathbf{V}\}$ in some of our simplified cases.

\begin{table}[!t]
\caption{Review of the coverage of \citep{Singh2021-og} across different prediction tasks including both $A$ and $S$. We refer to equalized odds for the anti-causal predictive case, and demographic parity for the causal predictive case, as in \citep{Veitch2021-rq}. Here, $\{\mathbf{V}\}$ is the set of variables to include as inputs to the model.}
\label{tab:limitations_featsel}       
\centering
\begin{tabular}{p{3cm}|p{2.5cm}|p{1cm}}
\hline
Predictive task & Shift & $\{\mathbf{V}\}$ \\
\hline
Anti-causal & demographic  & $\{X\}$ \\
 & covariate & $\{A\}$ \\
 & label & $\emptyset$ \\
 & compound & $\emptyset$ \\
\hline
Causal & demographic  & $\{X,A\}$ \\
 & covariate & $\emptyset$ \\
 & label & $\emptyset$ \\
 & compound & $\emptyset$ \\
\hline
\end{tabular}
\end{table}

\section{Data and method availability}
\label{app:data_avail}

The dermatology data is not available to the public. The de-identified EHR data is available based on a user agreement at Physionet \citep{Goldberger2000-zg}. The code for extracting Elixhauser and van Walren comorbidity scores from MIMIC-III is available at \url{https://doi.org/10.5281/zenodo.821872} \citep{Johnson2018-zj}. We take inspiration from the code made publicly by the authors of \citep{Tomasev2021-uf, Tomasev2021-uf} and available at \url{https://github.com/google/ehr-predictions}. We use scikit-learn \citep{scikit-learn} to estimate the balancing weights.

\section{Computational resources}
\label{app:comp_resources}

Our statistical tests include two main components:
\begin{itemize}
    \item Estimation of $P(S \mid \{\mathbf{V}\})$: logistic regression or boosted gradient trees are trained 100 times (number of bootstrap samples) for each test. At a maximum, this operation uses 12 Gb of RAM and 2 Gb of memory. All such model trainings are performed in notebooks.
    \item Summary of $U$: when the dimensionality $l$ of $U$ is larger than a couple of dozens, we summarize $U$ by training a model $f(U) \rightarrow Y$. For dermatology, a model training and tuning requires 28 TPU resources over 24-30 hours. For EHR, a model training requires 2 CPU for $\sim$1 hour.
\end{itemize}

To assess model performance and fairness across multiple distribution shifts, we train 10 replicates of a dermatology model $f(X,A) \rightarrow Y$ on the source data, and 10 replicates of an EHR model $f(M,X,T) \rightarrow Y$. We also train 10 replicates for the joint training in dermatology.

The training of larger models is performed on an internal cluster.

\section{Dermatology}

\subsection{Ethics approval and data availability}
\label{app:derm_data_avail}
The images and metadata were de-identified according to Health Insurance Portability and Accountability Act (HIPAA) Safe Harbor prior to transfer to study investigators. The protocol was reviewed by Advarra IRB (Columbia, MD), which determined that it was exempt from further review under 45 CFR 46. The dermatology data is not available to the public.

\subsection{Model and performance}

\subsubsection{Model architecture}
We train 10 replicates from different random seeds of a deep learning model to predict skin conditions from the images $X$, age and sex, with an approach similar to \citep{Liu2020-qr,Roy2021-xp}. To train the model, we consider all images pertaining to a case (min 1, max 6). Each image is resized to $448 \times 448$ pixels and encoded with a wide ResNet-101$\times$3 feature extractor initialised using BiT-L pretraining checkpoints \citep{bit}. The embeddings are then averaged across images and concatenated with the metadata (here age and sex) before passing through a fully connected layer which is followed by classification heads predicting the 26 + 1 conditions \citep{Liu2020-qr,Roy2021-xp}. An additional classification head covering a more fine-grained set of 419 conditions, defined on the same examples, is used at train time only.  

\subsubsection{Fairness properties are affected by the environment}
\label{app:derm_target1_model_perf}
\begin{figure*}[!t]
\centering
\includegraphics[width=0.7\linewidth]{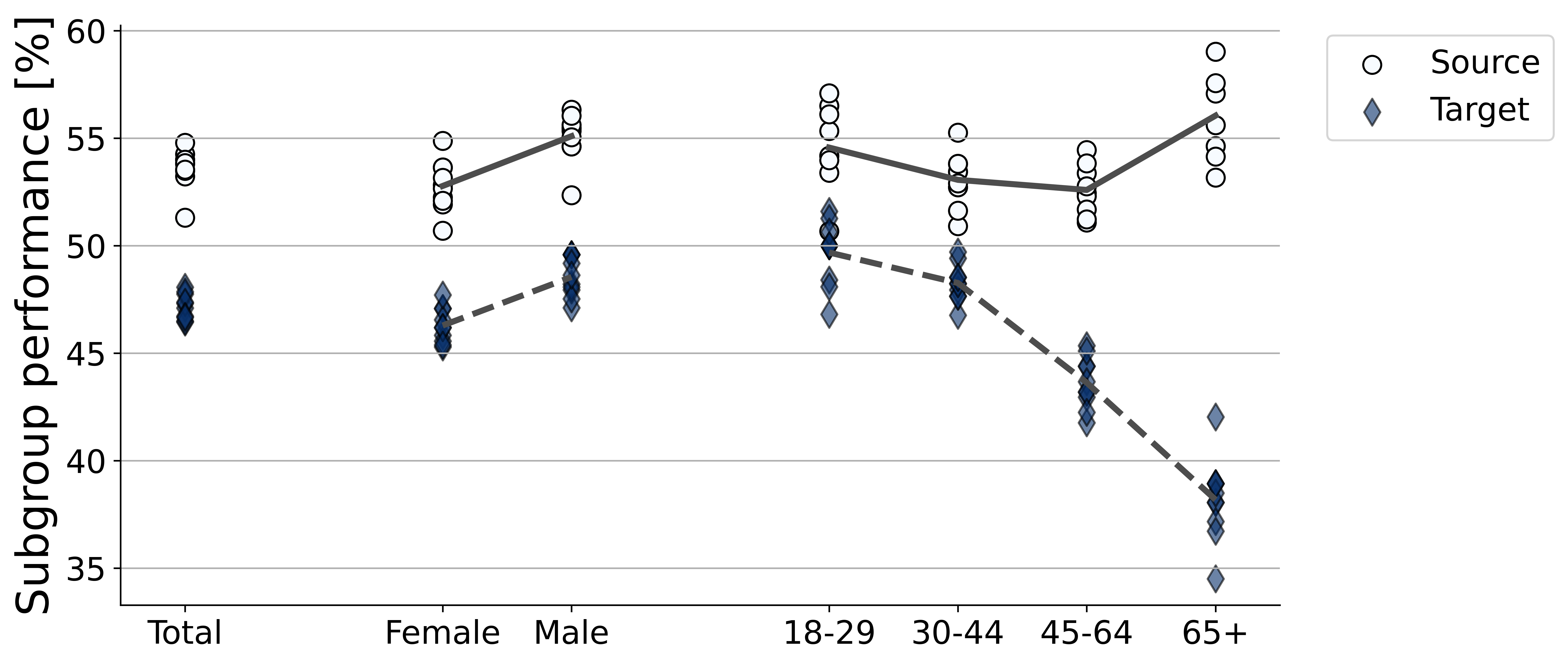} 
\caption{Model performance in dermatology, as estimated via Top-1 accuracy (in \%). The plot displays the total performance, as well as performance stratified by sex and by age on the source (circles with plain line) and target (diamonds with dashed line) data. Each marker represents one replicate of the model.}
\label{fig:derm_top1_target1}       
\end{figure*}

Detailed top-3 and top-1 model performance can be found in Tables \ref{tab:top3_derm_target1} and \ref{tab:top1_derm_target1}, respectively. For top-1 accuracy, we observe similar results as for top-3 accuracy: model performance is relatively similar between groups on the source data, but differences become apparent on the target data (Figure~\ref{fig:derm_top1_target1}), especially for age. Interestingly, the target includes 544 paediatric cases, on which the model performs better than in other age groups (top-3: $85.28 \pm 1.32 \%$, top-1: $51.53 \pm 1.81 \%$).

\begin{table}[!h]
\centering
\caption{Top-3 model accuracy (in \%) in the source and target data, on average across model runs.}
\label{tab:top3_derm_target1}       
\begin{tabular}{lll}
\hline\noalign{\smallskip}
Group & Source & Target  \\
\noalign{\smallskip}\hline\noalign{\smallskip}
Total & $88.52 \pm 0.68$ & $79.35 \pm 1.02$ \\
\noalign{\smallskip}\hline\noalign{\smallskip}
Female & $88.95 \pm 0.93$ (n=1,221) & $79.02 \pm 0.92$ (n=1,115) \\
Male & $87.78 \pm 0.52$ (n=703) & $79.85 \pm 1.28$ (n=728) \\
\noalign{\smallskip}\hline\noalign{\smallskip}
$[18, 30)$ & $88.51 \pm 1.11$ (n=563) & $81.85 \pm 1.74$ (n=314) \\
$[30, 45)$ & $88.38 \pm 1.02$ (n=548) & $80.21 \pm 0.93$ (n=340) \\
$[45, 65)$ & $88.89 \pm 0.70$ (n=619) & $74.68 \pm 1.45$ (n=419) \\
$[65, 90)$ & $87.84 \pm 1.70$ (n=194) & $68.98 \pm 2.00$ (n=226) \\
\noalign{\smallskip}\hline
\end{tabular}
\end{table}

\begin{table}[!h]
\centering
\caption{Top-1 model accuracy (in \%) in the source and target data, on average across model runs.}
\label{tab:top1_derm_target1}       
\begin{tabular}{lll}
\hline\noalign{\smallskip}
Group & Source & Target  \\
\noalign{\smallskip}\hline\noalign{\smallskip}
Total & $53.62 \pm 0.88$ & $47.18 \pm 0.56$ \\
\noalign{\smallskip}\hline\noalign{\smallskip}
Female & $53.62 \pm 0.88$ & $46.29 \pm 0.77$ \\
Male & $55.09 \pm 1.05$ & $48.54 \pm 0.86$ \\
\noalign{\smallskip}\hline\noalign{\smallskip}
$[18, 30)$ & $54.49 \pm 1.73$ & $49.68 \pm 1.41$ \\
$[30, 45)$ & $53.12 \pm 1.14$ & $48.26 \pm 0.82$ \\
$[45, 65)$ & $52.68 \pm 0.92$ & $43.63 \pm 1.12$ \\
$[65, 90)$ & $55.52 \pm 2.00$ & $38.19 \pm 1.83$ \\
\noalign{\smallskip}\hline
\end{tabular}
\end{table}

\subsubsection{Per condition model performance}
\label{app:derm_per_cond_target1}

\begin{figure*}[!ht]
\centering
\begin{subfigure}[t]{0.03\textwidth}
(a)
\end{subfigure}
\begin{subfigure}[t]{0.6\textwidth}
\includegraphics[width=\linewidth,valign=t]{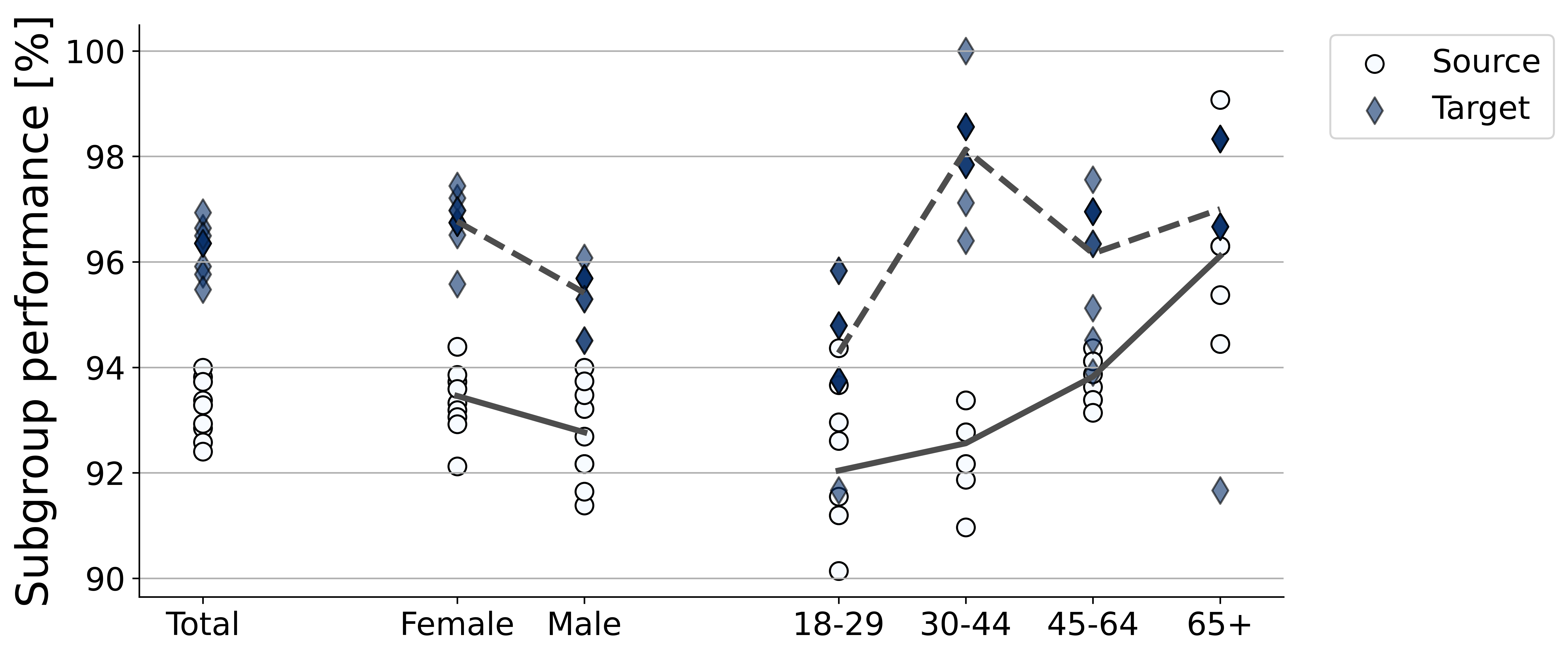} 
\end{subfigure}\\
\begin{subfigure}[t]{0.03\textwidth}
(b)
\end{subfigure}
\begin{subfigure}[t]{0.6\textwidth}
\includegraphics[width=\linewidth,valign=t]{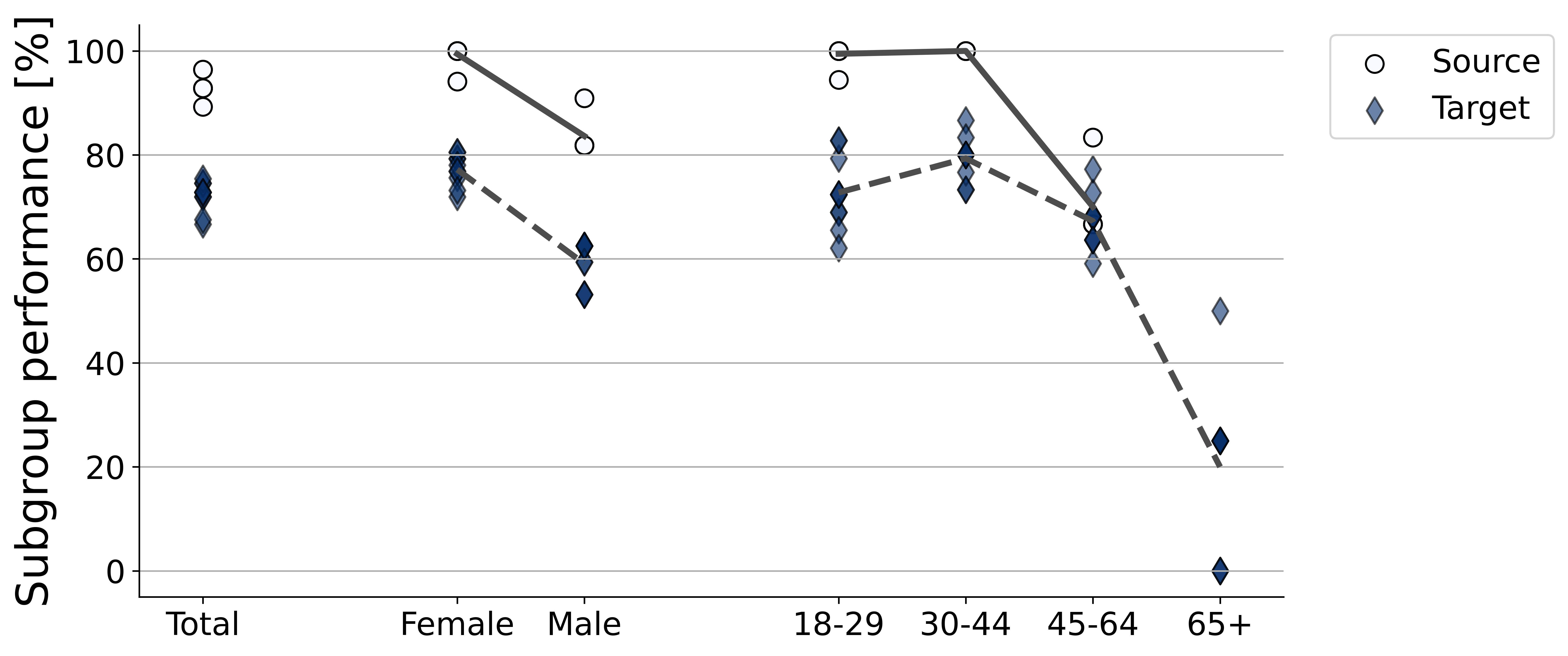}
\end{subfigure}\\
\begin{subfigure}[t]{0.03\textwidth}
(c)
\end{subfigure}
\begin{subfigure}[t]{0.6\textwidth}
\includegraphics[width=\linewidth,valign=t]{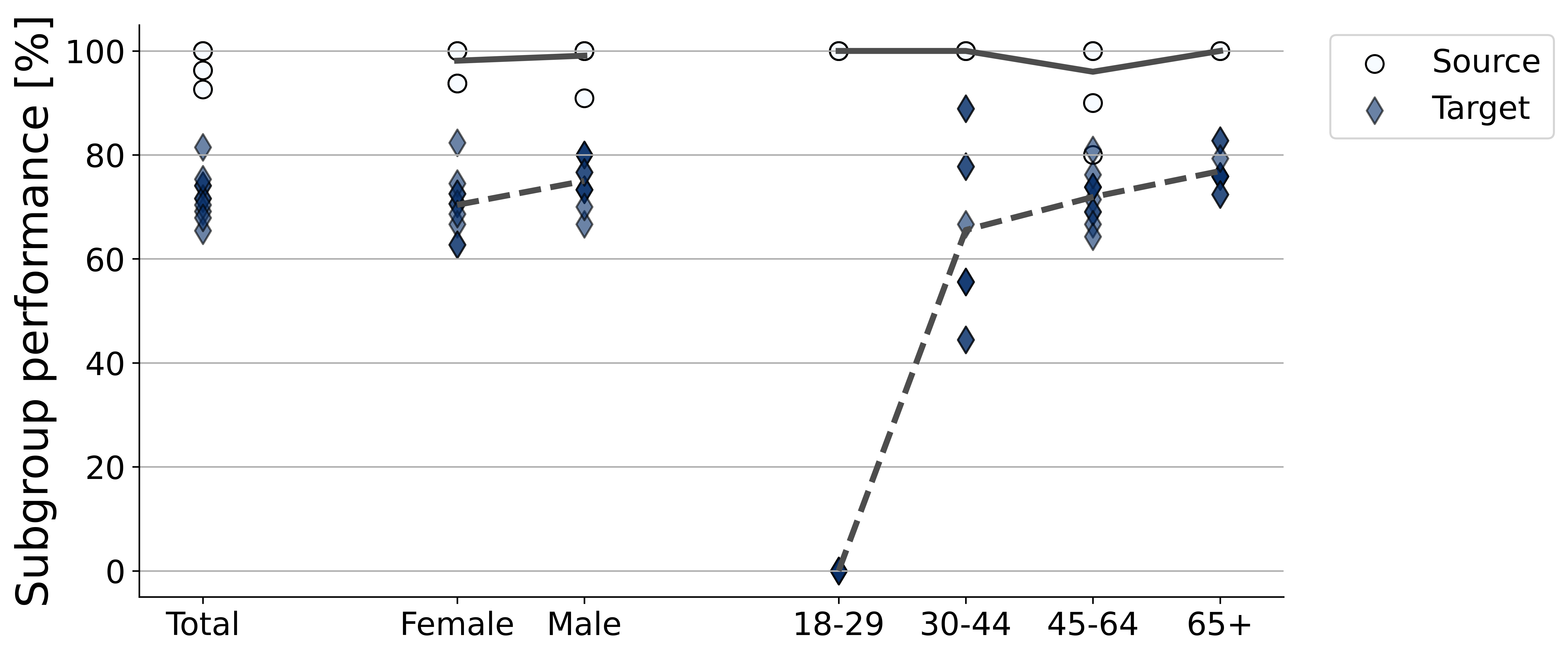}
\end{subfigure}
\caption{Per condition model performance in dermatology, as estimated by top-3 accuracy (in \%). (a) 'Other'. (b) Melanocytic Nevus. (c) SK/ISK.}
\label{fig:derm_top3_per_condition_aegir}       
\end{figure*}

We further compute model performance per condition when enough data samples are available both in the source and target data. We first observe that the tail of other conditions (`other', Figure~\ref{fig:derm_top3_per_condition_aegir}(a)) shows more similarity with the source, and even improved performance. The fairness patterns for Melanocytic Nevus also seem to be similar across datasets, decreasing with age. On the other hand, SK/ISK shows an increase in performance with age in the target, but not in the source. This last result highlights that the changes in fairness patterns do not only result from changes in condition prevalence. We note that the analyses in this section include relatively low number of samples per subgroup.

\subsubsection{Fitzpatrick's skin type}
\label{app:derm_target2_skintype}
\begin{figure*}[!ht]
\centering
\begin{subfigure}[t]{0.03\textwidth}
(a)
\end{subfigure}
\begin{subfigure}[t]{0.33\textwidth}
\includegraphics[width=\linewidth,valign=t]{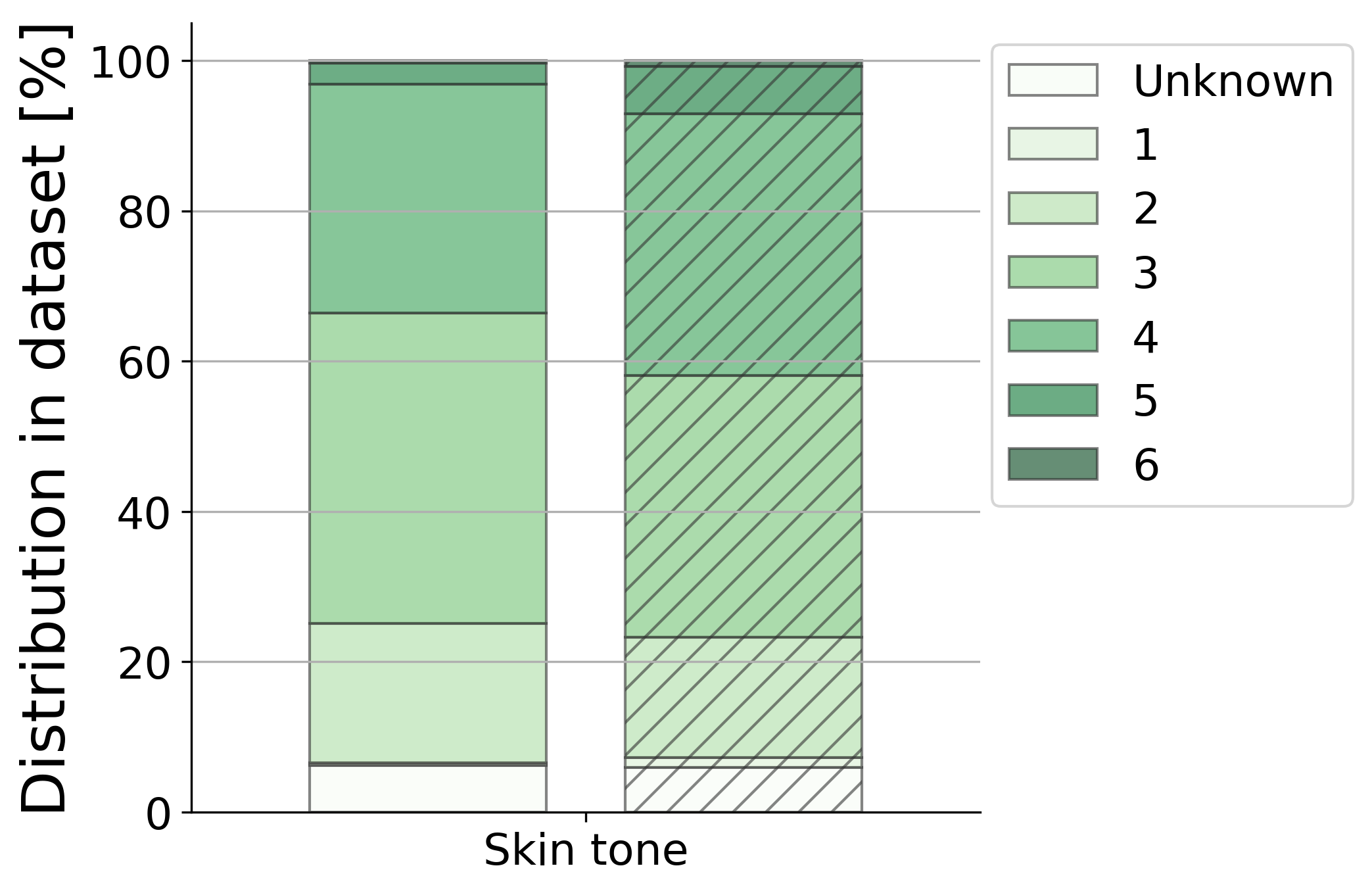} 
\end{subfigure}
\begin{subfigure}[t]{0.03\textwidth}
(b)
\end{subfigure}
\begin{subfigure}[t]{0.57\textwidth}
\includegraphics[width=\linewidth,valign=t]{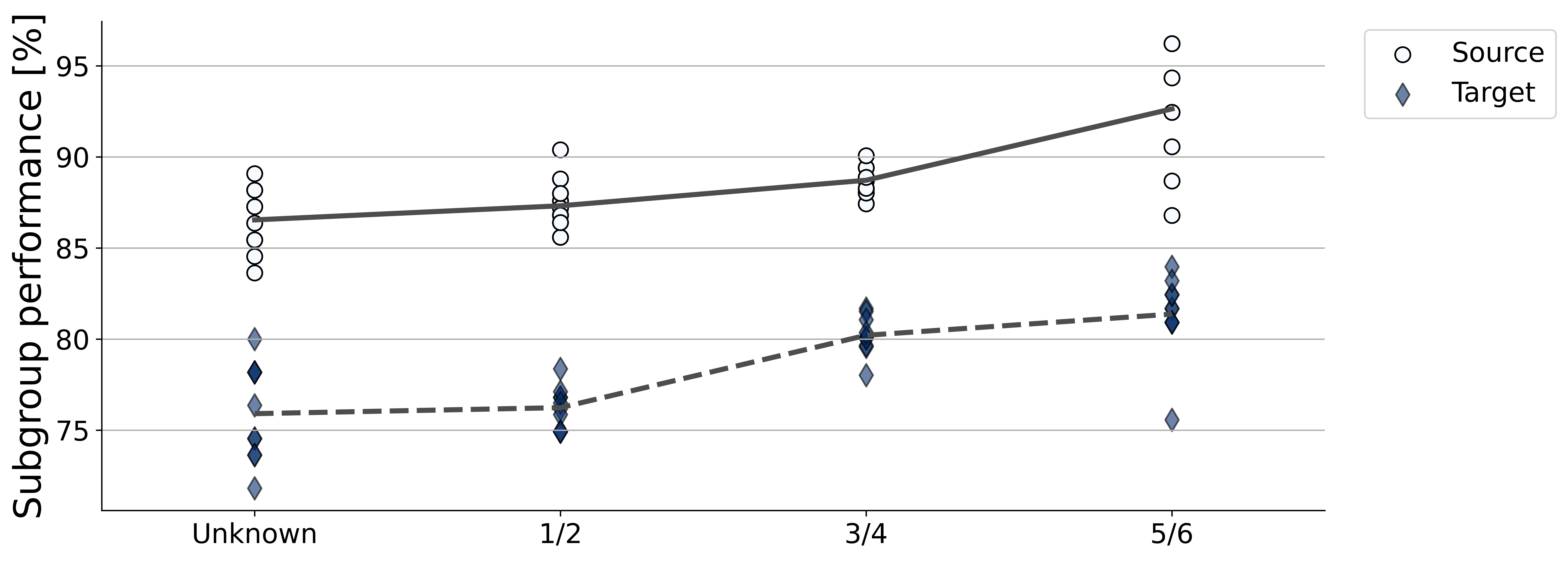}
\end{subfigure}
\caption{(a) Prevalence of the six Fitzpatrick's skin types in each dataset (left: source, right: target). (b) Top-3 model performance for each subgroup in the source and in the target.}
\label{fig:derm_skintone}       
\end{figure*}

In this target dataset, skin type is available as an attribute. We therefore perform similar analyses as performed for age and sex. We first observe that the proportions of cases across the different skin types are different across the source and target datasets (Figure~\ref{fig:derm_skintone}(a), t-test: $p<0.001$).

Due to a low number of examples for skin types 1 and 6 (in both datasets), we group skin types according to `unknown', `1/2', `3/4' and `5/6' before assessing top-3 performance per subgroup. We observe that model performance is relatively similar across the different subgroups, with the maximum pairwise difference being 6\% on the source and 5.5\% on the target (Figure~\ref{fig:derm_skintone}(b)). We however note that, despite the groupings, the number of examples in group 5/6 remains low (source: $n=53$, target: $n=131$). Performance results in this group are hence to take with a grain of salt (as displayed by large variability across seeds), and single conditions cannot be investigated.

\subsection{Second target dataset: skin cancer clinics in Australia}
\label{app:derm_target2}
In this section, we repeat the analyses performed in the main text for another target dataset from skin cancer clinics in Australia and New Zealand that was partially labelled, leading to 21,661 cases used for model evaluation.

\subsubsection{Fairness properties are affected by the environment}
\label{app:derm_target2_model_perf}
\begin{figure*}[!t]
\centering
\includegraphics[width=0.7\linewidth]{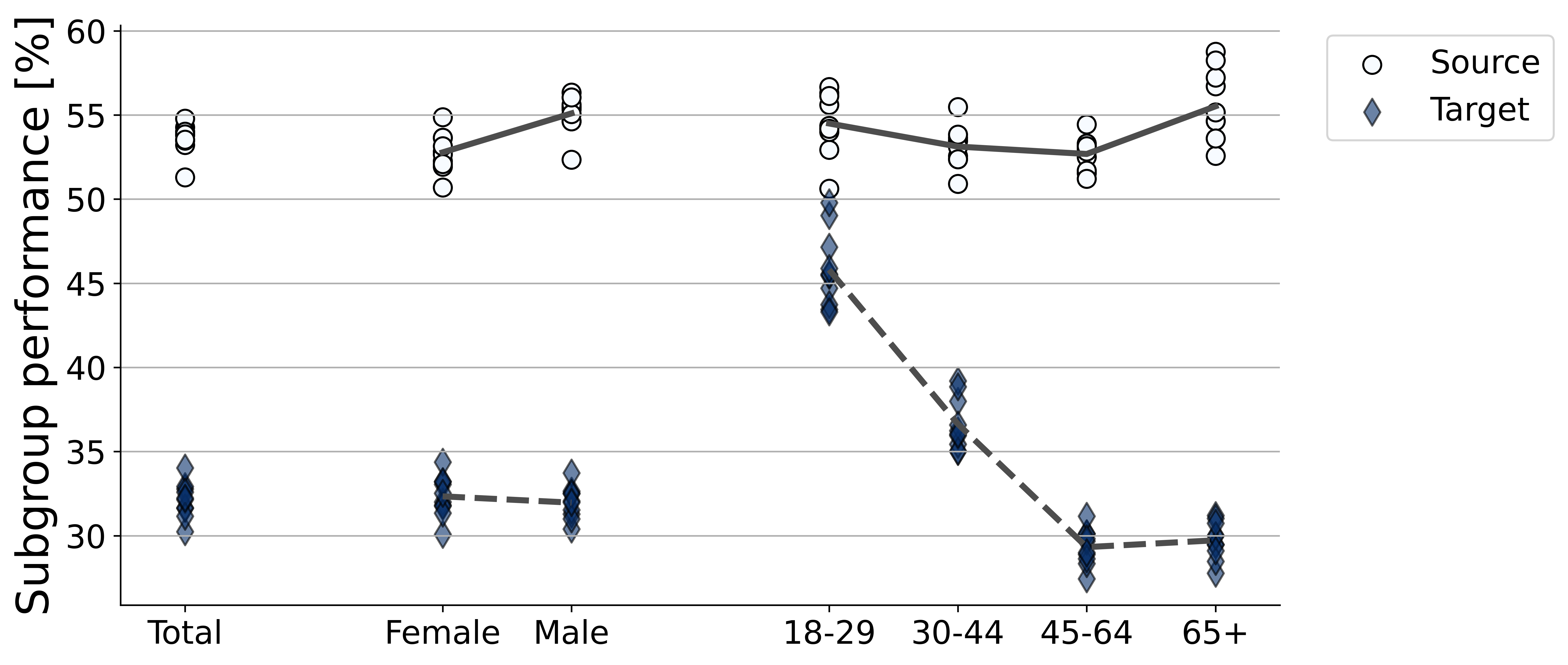} 
\caption{Model performance in dermatology, as estimated via Top-1 accuracy (in \%). The plot displays the total performance, as well as performance stratified by sex and by age on the source (circles with plain line) and target (diamonds with dashed line) data. Each marker represents one replicate of the model.}
\label{fig:derm_top1_target2}       
\end{figure*}

Detailed top-3 and top-1 model performance can be found in Tables \ref{tab:top3_derm_target2} and \ref{tab:top1_derm_target2}, respectively. For top-1 accuracy, we observe similar results as for top-3 accuracy: model performance is relatively similar between groups on the source data, but differences become apparent on the target data (Figure~\ref{fig:derm_top1_target2}).

\begin{table}[!h]
\centering
\caption{Top-3 model accuracy (in \%) in the source and target data, on average across model runs.}
\label{tab:top3_derm_target2}       
\begin{tabular}{lll}
\hline\noalign{\smallskip}
Group & Source & Target  \\
\noalign{\smallskip}\hline\noalign{\smallskip}
Total & $88.52 \pm 0.68$ & $70.87 \pm 0.85$ \\
\noalign{\smallskip}\hline\noalign{\smallskip}
Female & $88.95 \pm 0.93$ (n=1,221) & $72.11 \pm 0.98$ (n=10,195) \\
Male & $87.78 \pm 0.52$ (n=703) & $69.77 \pm 0.80$ (n=11,466) \\
\noalign{\smallskip}\hline\noalign{\smallskip}
$[18, 30)$ & $88.51 \pm 1.11$ (n=563) & $87.52 \pm 1.15$ (n=1,434) \\
$[30, 45)$ & $88.38 \pm 1.02$ (n=548) & $77.64 \pm 0.79$ (n=4,365) \\
$[45, 65)$ & $88.89 \pm 0.70$ (n=619) & $68.39 \pm 0.97$ (n=8,355) \\
$[65, 90)$ & $87.84 \pm 1.70$ (n=194) & $66.20 \pm 1.44$ (n=7,401) \\
\noalign{\smallskip}\hline
\end{tabular}
\end{table}

\begin{table}[!h]
\centering
\caption{Top-1 model accuracy (in \%) in the source and target data, on average across model runs.}
\label{tab:top1_derm_target2}       
\begin{tabular}{lll}
\hline\noalign{\smallskip}
Group & Source & Target  \\
\noalign{\smallskip}\hline\noalign{\smallskip}
Total & $53.62 \pm 0.88$ & $32.14 \pm 0.99$ \\
\noalign{\smallskip}\hline\noalign{\smallskip}
Female & $53.62 \pm 0.88$ & $32.34 \pm 1.14$ \\
Male & $55.09 \pm 1.05$ & $31.96 \pm 0.91$ \\
\noalign{\smallskip}\hline\noalign{\smallskip}
$[18, 30)$ & $54.49 \pm 1.73$ & $45.80 \pm 2.14$ \\
$[30, 45)$ & $53.12 \pm 1.14$ & $36.63 \pm 1.45$ \\
$[45, 65)$ & $52.68 \pm 0.92$ & $29.32 \pm 1.01$ \\
$[65, 90)$ & $55.52 \pm 2.00$ & $29.72 \pm 1.05$ \\
\noalign{\smallskip}\hline
\end{tabular}
\end{table}

\subsubsection{Per condition model performance}
\label{app:derm_per_cond_target2}

\begin{figure*}[!ht]
\centering
\begin{subfigure}[t]{0.03\textwidth}
(a)
\end{subfigure}
\begin{subfigure}[t]{0.6\textwidth}
\includegraphics[width=\linewidth,valign=t]{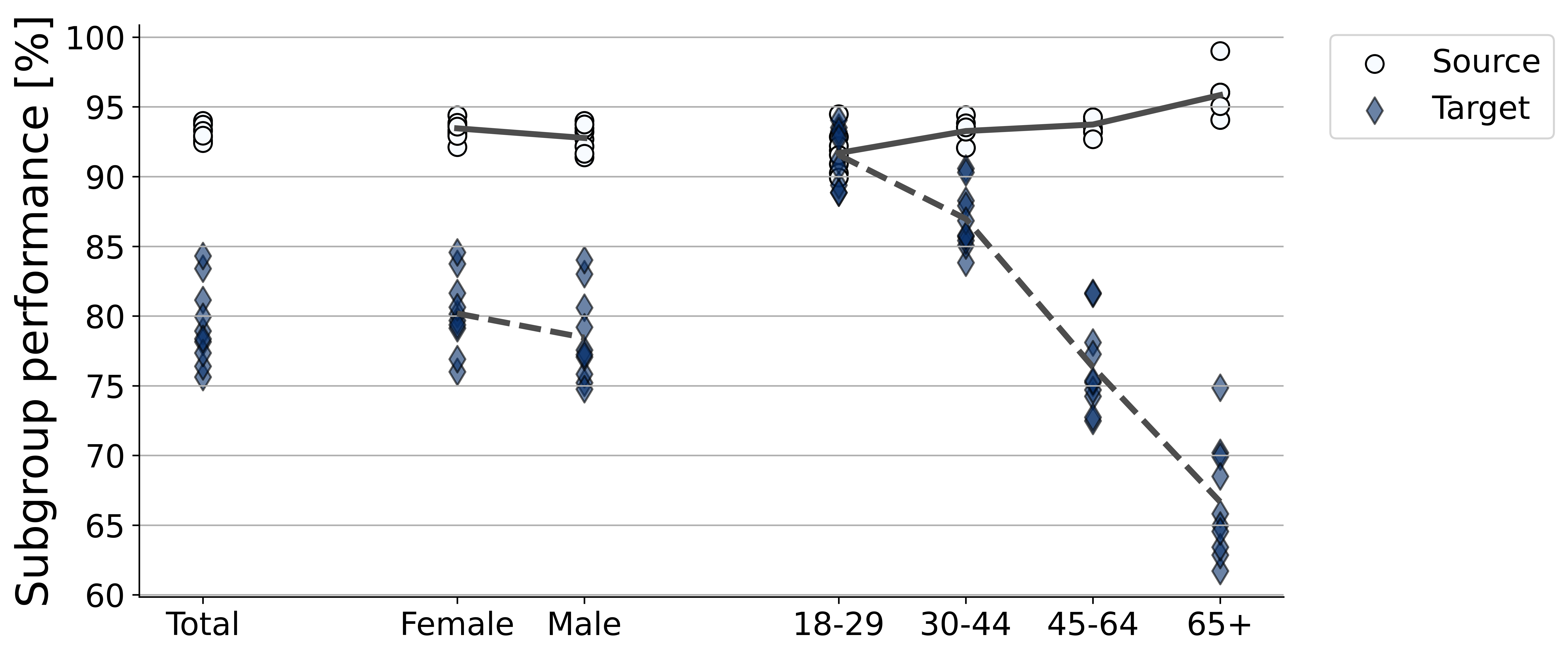} 
\end{subfigure}\\
\begin{subfigure}[t]{0.03\textwidth}
(b)
\end{subfigure}
\begin{subfigure}[t]{0.6\textwidth}
\includegraphics[width=\linewidth,valign=t]{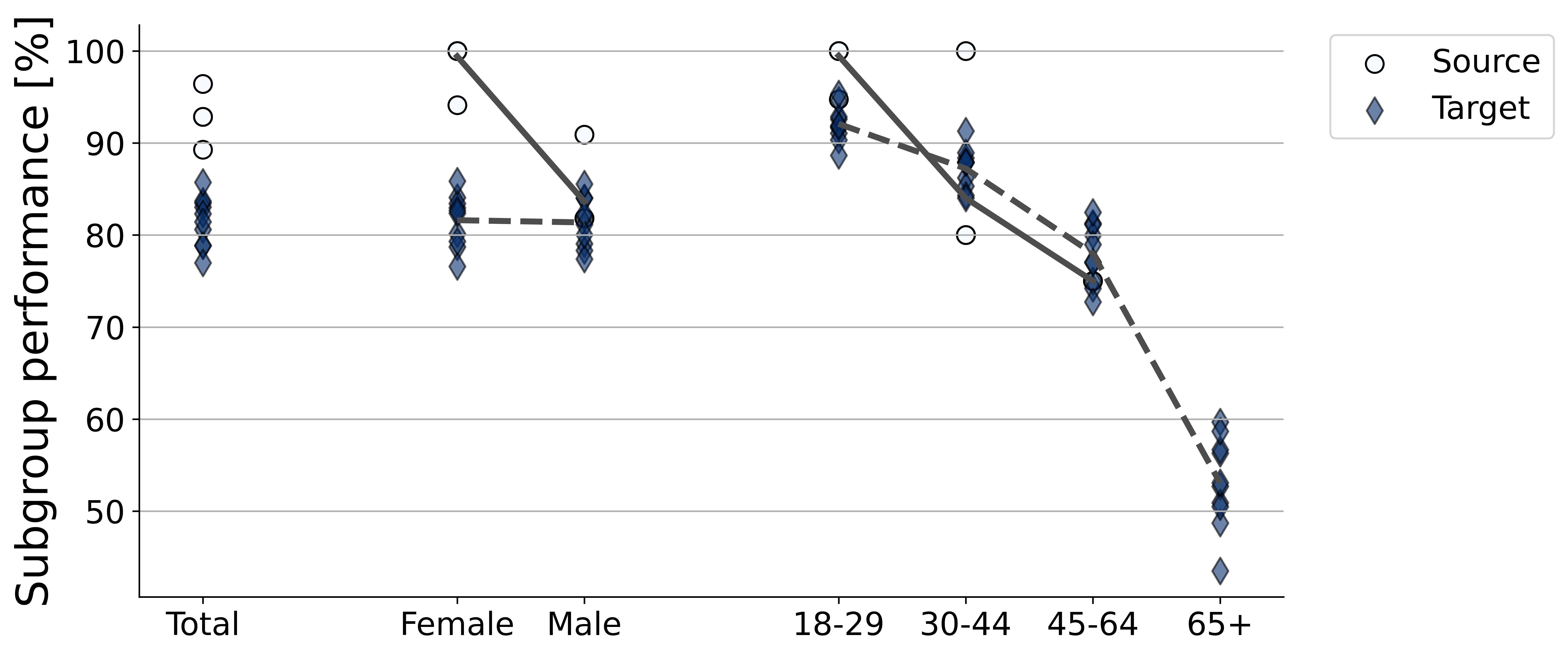}
\end{subfigure}\\
\begin{subfigure}[t]{0.03\textwidth}
(c)
\end{subfigure}
\begin{subfigure}[t]{0.6\textwidth}
\includegraphics[width=\linewidth,valign=t]{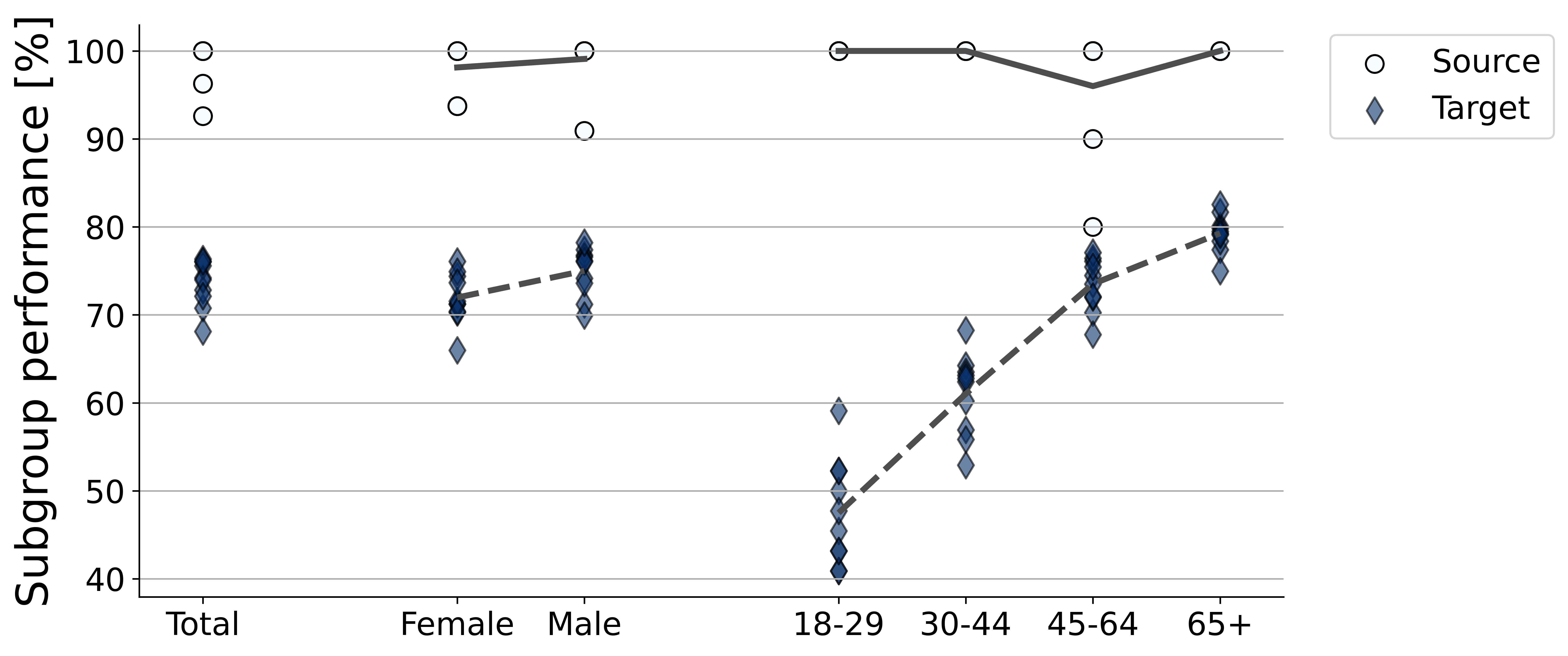}
\end{subfigure}
\caption{Per condition model performance in dermatology, as estimated by top-3 accuracy (in \%). (a) 'Other'. (b) Melanocytic Nevus. (c) SK/ISK.}
\label{fig:derm_top3_per_condition_target2}       
\end{figure*}

 We repeat the analyses for the same conditions depicted in Section~\ref{app:derm_per_cond_target1}. We observe that the gaps between groups vary based on the condition, with the long tail of conditions (represented as a single `other' condition in our task) being seemingly `fair' on the source, but not on the target (Figure~\ref{fig:derm_top3_per_condition_target2}(a)). Some conditions like Melanocytic Nevus present disparities across age groups in both datasets (with the caveat of few test samples in the source data), with performance decreasing with increasing age (max gap in source: 24.47\%, in target: 39.00\%, Figure~\ref{fig:derm_top3_per_condition_target2}(b)). On the other hand, the SK/ISK condition displays similar performance across age groups in the source, but increasing performance with increasing age in the target (max gap in source: 4.00\%, max gap in target: 31.78\%, Figure~\ref{fig:derm_top3_per_condition_target2}(c)). Where the sample size allowed for comparisons, we therefore observe that the gaps across groups are not reproducible between the source and target data.

\subsubsection{A compound shift}
Referring to the same causal graph as in the main text, we apply our testing procedure to this second target set.

\begin{figure*}[!t]
\centering
\begin{subfigure}[t]{0.03\textwidth}
(a)
\end{subfigure}
\begin{subfigure}[t]{0.45\textwidth}
\resizebox{0.7\linewidth}{!}{
\begin{tikzpicture}[baseline={([yshift={-\ht\strutbox}]current bounding box.north)},outer sep=0pt,inner sep=0pt]
\node (S) at (-0.25,-3) {$S$};
\node (X) at (1,-4) {$X$};
\node (A) at (-1.5,-4) {$A$};
\node (Y) at (-0.25,-4.5) {$Y$};
\draw[line width=1pt,orange, \arr, opacity=0.6](S)--(Y);
\draw[line width=1pt,blue, \arr, opacity=0.6](S)--(A);
\draw[line width=1pt,green!40!black, \arr, opacity=0.6](S)--(X);
\draw[line width=1pt,black, \arr, opacity=0.6](A)--(X);
\draw[line width=1pt,black, \arr, opacity=0.7](A)--(Y);
\draw[line width=1pt,black, \arr, opacity=0.7](Y)--(X);
\end{tikzpicture}}
\end{subfigure}
\begin{subfigure}[t]{0.03\textwidth}
(b)
\end{subfigure}
\begin{subfigure}[t]{0.45\textwidth}
\includegraphics[width=0.9\linewidth,valign=t]{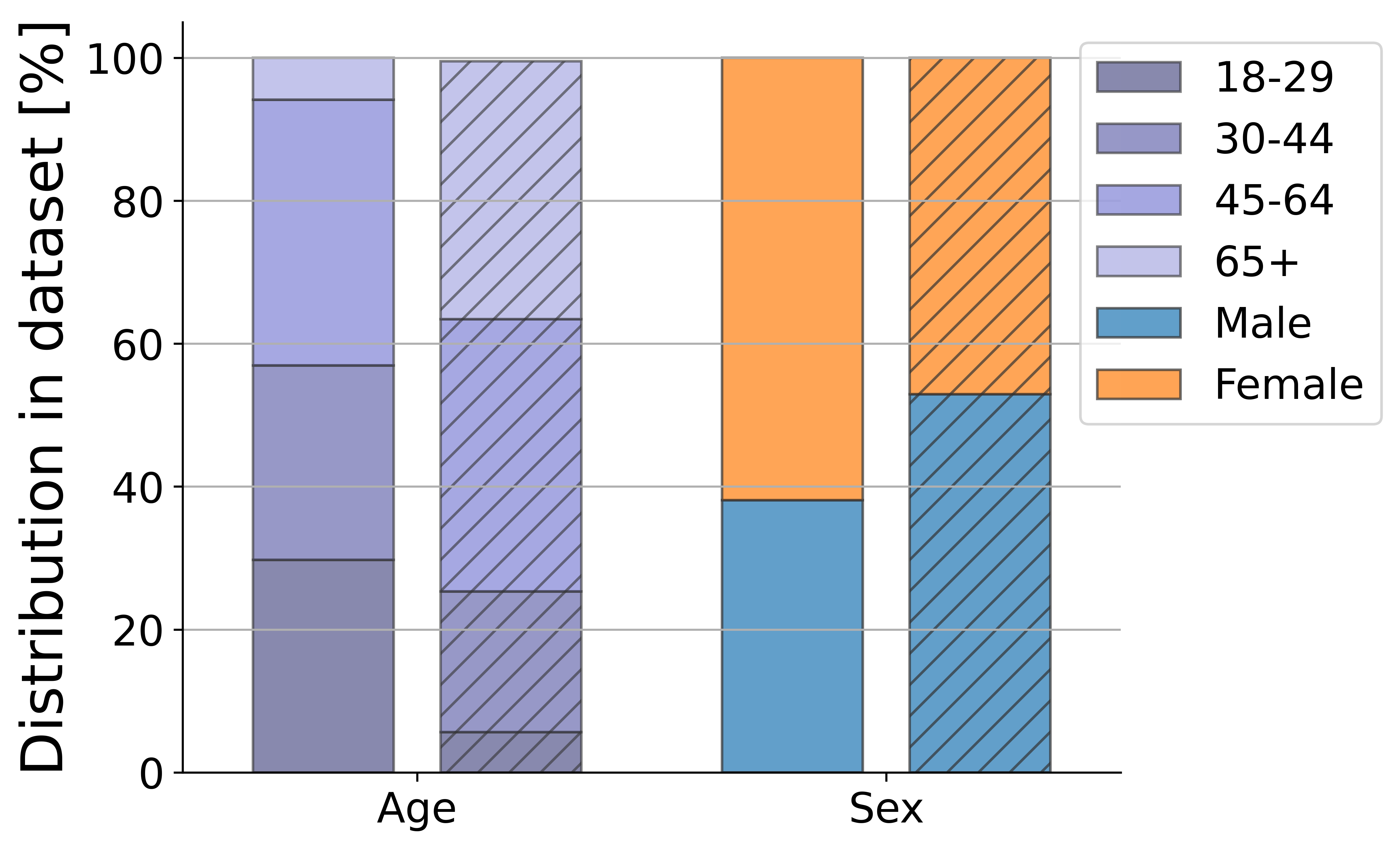} 
\end{subfigure}\\
\begin{subfigure}[t]{0.03\textwidth}
(c)
\end{subfigure}
\begin{subfigure}[t]{0.55\textwidth}
\includegraphics[width=\linewidth,valign=t]{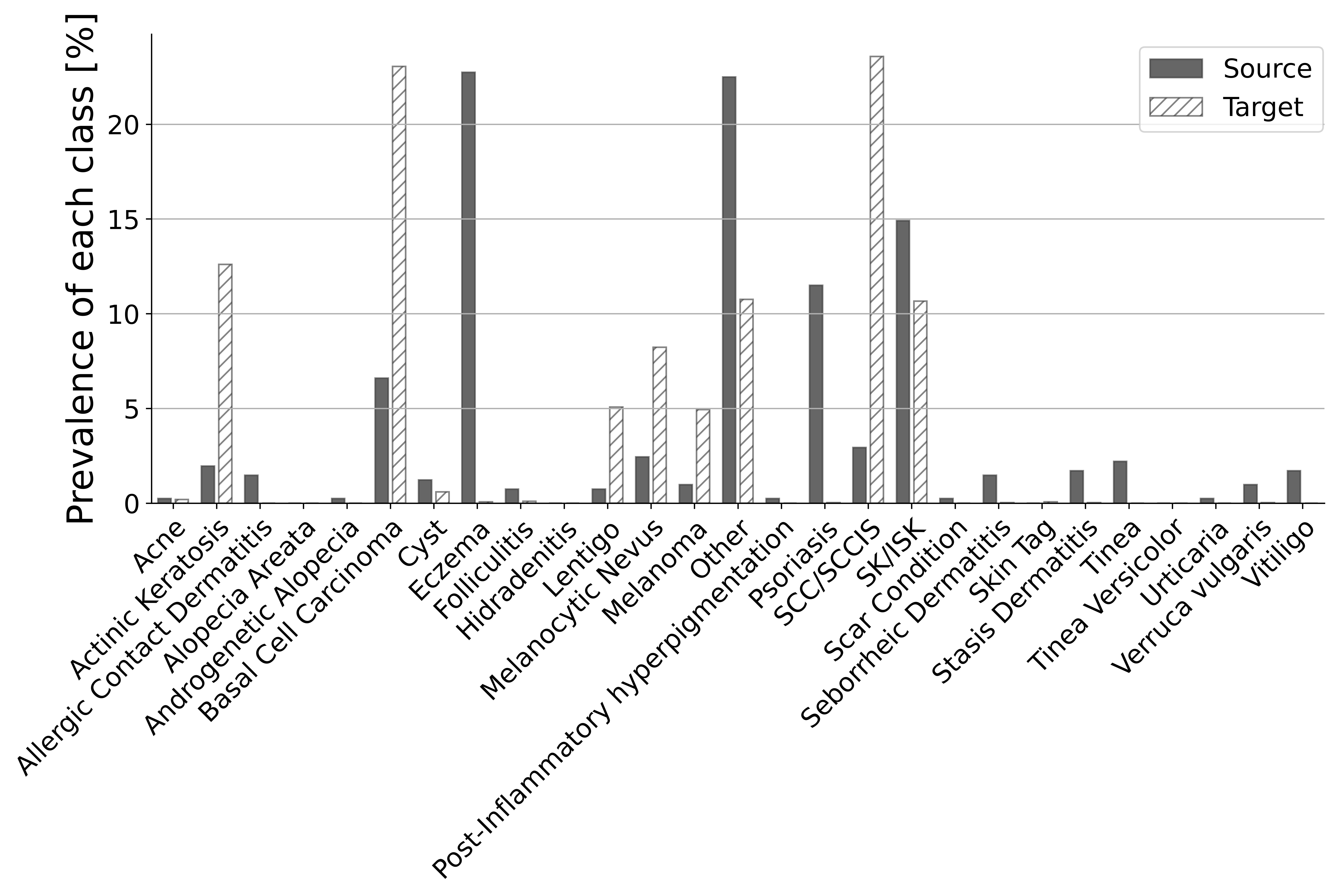}
\end{subfigure}
\begin{subfigure}[t]{0.03\textwidth}
(d)
\end{subfigure}
\begin{subfigure}[t]{0.33\textwidth}
\includegraphics[width=\linewidth,valign=t]{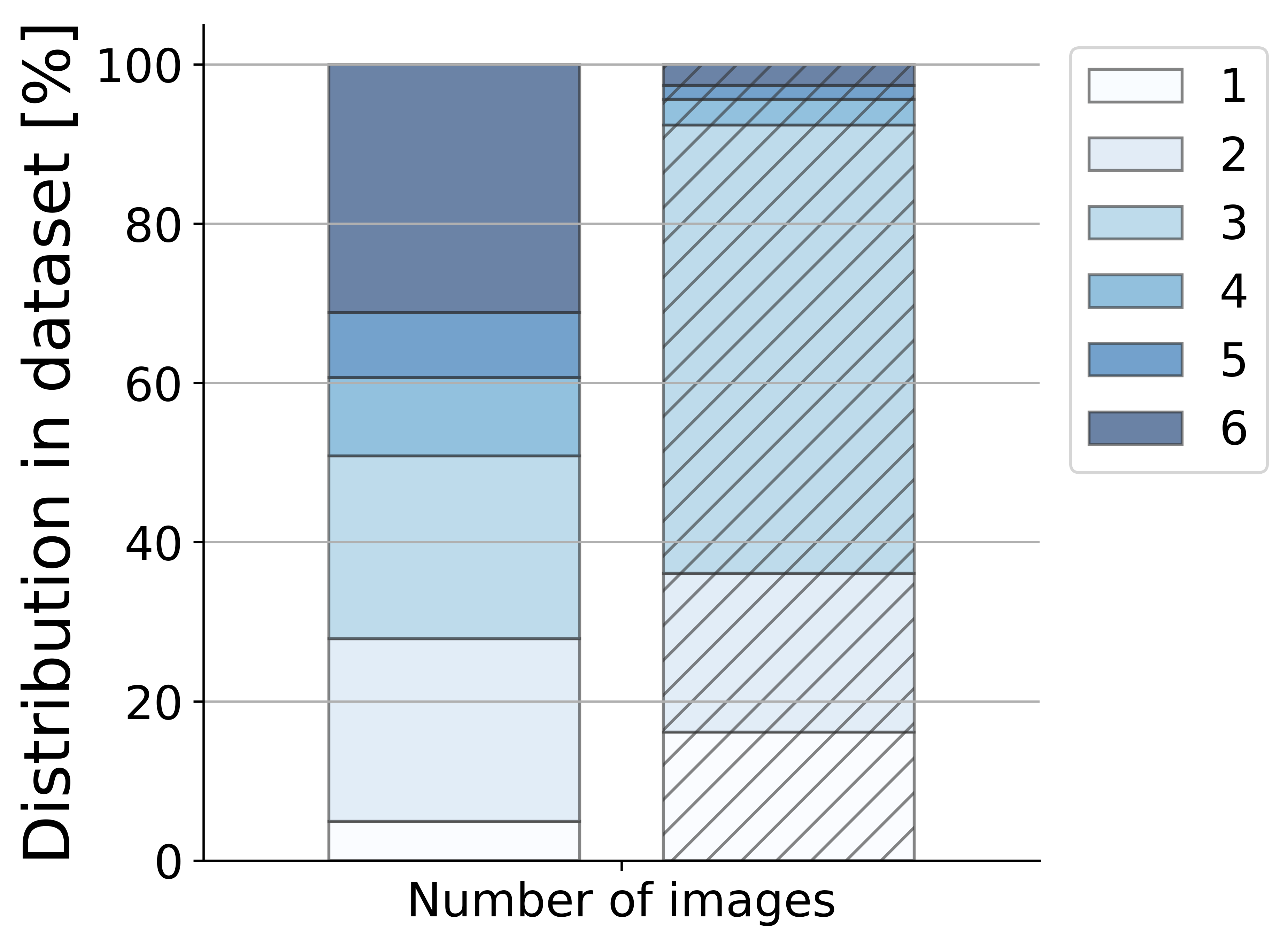}
\end{subfigure}
\caption{A compound shift in dermatology. (a) Simplified causal structure of the application. $A$ refers to demographic metadata, $Y$ to the skin condition as labelled by clinicians and $X$ represents the set of pictures of the skin pathology. In our setup, the environment $S$ directly affects all variables in the graph. (b) Distribution in the source train and in the target dataset of the sensitive attributes, computed in terms of percentage of pre-defined subgroups. (c) Prevalence of each condition in cases where the patients are female and over 65 years of age. (d) Distribution of the number of images in cases labelled as `SK/ISK' in older females. The distributions in the source data are represented on the left, and their corresponding distributions in the target data on their immediate right with a hashed pattern.}
\label{fig:derm_shift}       
\end{figure*}

Only age and sex\footnote{Sex is mainly recorded by clinicians, with a small set of the source data containing self-reported sex.} are available in both datasets. These features have different distributions across the two datasets (see Fig.~\ref{fig:derm_shift}(b)), with the population in the source data being typically younger (median age: 40 years old, 25$\%$ quantile: 27, 75$\%$ quantile: 54) than in the target data (median age: 58 years old, 25$\%$ quantile: 44, 75$\%$ quantile: 70). The source distribution also includes more female patients than the target data (62$\%$, compared to 47$\%$). We therefore see direct effects of $S$ on $A$ (Fig.~\ref{fig:derm_shift}a, in purple).

We then assess whether $S$ directly affects the labels. Based on our testing procedure ($\{\mathbf{V}\}=\{A\}$), we obtain significant differences between $P(Y \,|\, A, S=0)$ and $P(Y \,|\, A, S=1)$ for 25 conditions ($p<0.05$, Bonferroni corrected). Fig.\ref{fig:derm_shift}(c) illustrates the label shift in a specific age and sex subgroup (here females aged 65+, 409 cases in the source, 3,198 in the target data). We see that the source data includes more cases of `eczema' and `psoriasis' and that cancerous conditions such as `basal cell carcinoma' and `melanoma' are more prevalent in the target data. Our results suggest that the environment also directly affects the labels (orange link in Fig.\ref{fig:derm_shift}(a)). 
We note that this analysis is limited to the distribution of labels and cannot assess differences in their quality: for instance, the labels for the source data were sourced from multiple clinicians, while some target labels were verified by biopsies. This is an added difference between the two environments (known as annotation shift, \citep{castro2020causality}) that only in-depth dataset knowledge can bring.

Finally, we consider whether the features of the images themselves (designated by $X$) are directly affected by $S$. Our weighted tests suggest a significant difference between these two distributions ($p<0.05$ on 21 dimensions, corrected). Fig.\ref{fig:derm_shift}(d) illustrates this difference by computing the number of images per case in the group of older females considered above with cases labelled as `SK/ISK' ( source median: 3, $n=61$, target median: 2, $n=341$). This result suggests the existence of the direct path $S \rightarrow X$ and we add this relationship to the graph (green link in Fig.\ref{fig:derm_shift}(a)).

Based on these different analyses, we conclude that the environment is affecting all the variables in our simplified causal graph.

\subsection{Joint training}
\label{app:derm_joint_training}
In this section, we test whether joint training across datasets improves the transfer of fairness. We select the setup where the second target dataset (skin cancer clinics in Australia) is available for model development and add it to the source data, while we aim to obtain a robustly fair model on the first target dataset (teledermatology clinics in Colombia). This provides a training set that approximately doubles in size and now includes many more examples of cancer-related conditions. 

We first assess whether the distributions of this joint source more closely match the distribution of the target. To this end, we perform the same statistical tests as previously but select half of the source data from each dataset. This simulates an equal proportion of each dataset in the joint distribution.

We observe that $A$, $Y|A$ (5 conditions with $p<0.05$ corrected) and $X|A,Y$ (2 conditions with $p<0.05$ corrected) display effects of the environment. Interestingly, the joint source introduces an effect of $S$ on different dimensions than previously. For instance, we now notice a difference in the distribution of cancer-related conditions between the joint source and the target that was not noticeable when comparing the original source (teledermatology clinics in the US) to the target.

In terms of model training, we observe that combining the two datasets for joint training is non-trivial. This is due to the datasets being very different (see section \ref{app:derm_target2}). We take the simple approach of sampling a batch of each dataset with equal probability. This will oversample the data from the least represented dataset (here the original source). We note that this probability could be optimized over.

\begin{figure*}[!ht]
\centering
\begin{subfigure}[t]{0.03\textwidth}
\textbf{a}
\end{subfigure}
\begin{subfigure}[t]{0.6\textwidth}
\includegraphics[width=\linewidth,valign=t]{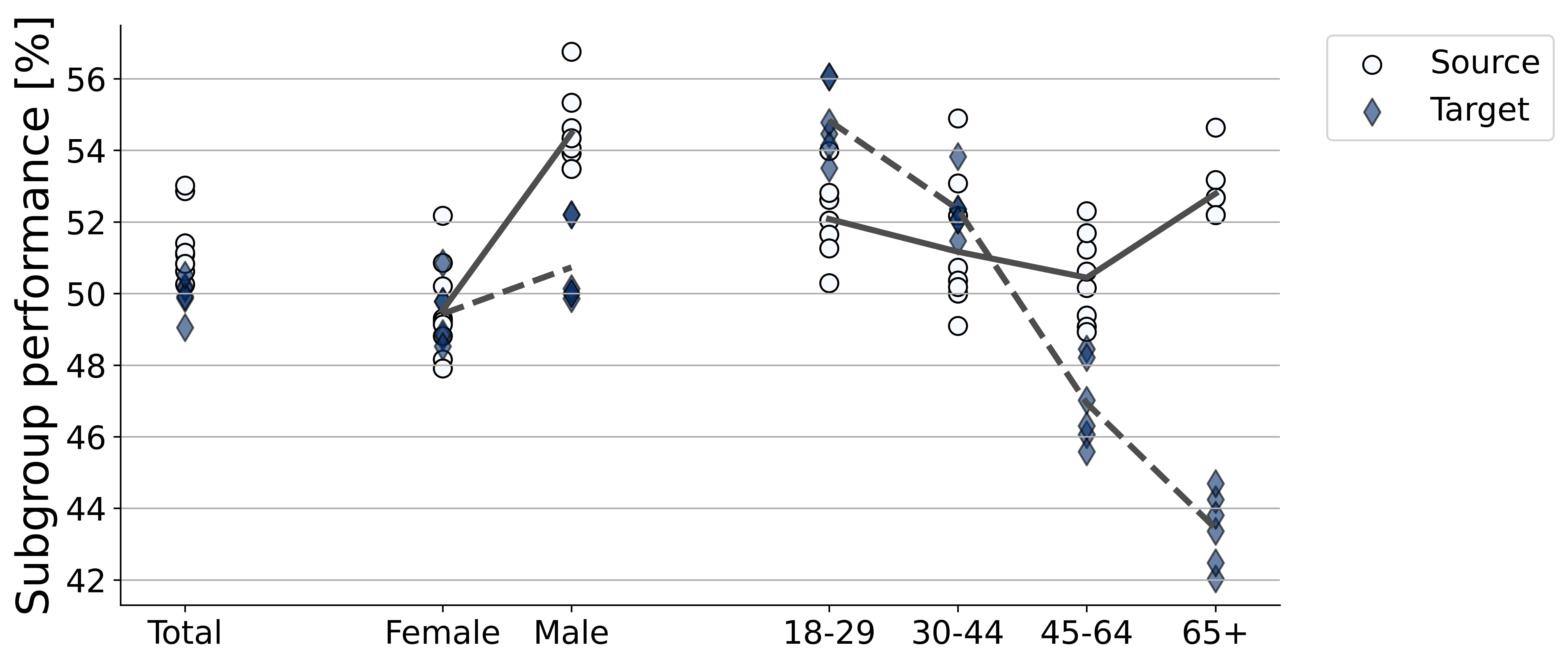}
\end{subfigure}\\
\begin{subfigure}[t]{0.03\textwidth}
\textbf{b}
\end{subfigure}
\begin{subfigure}[t]{0.6\textwidth}
\includegraphics[width=\linewidth,valign=t]{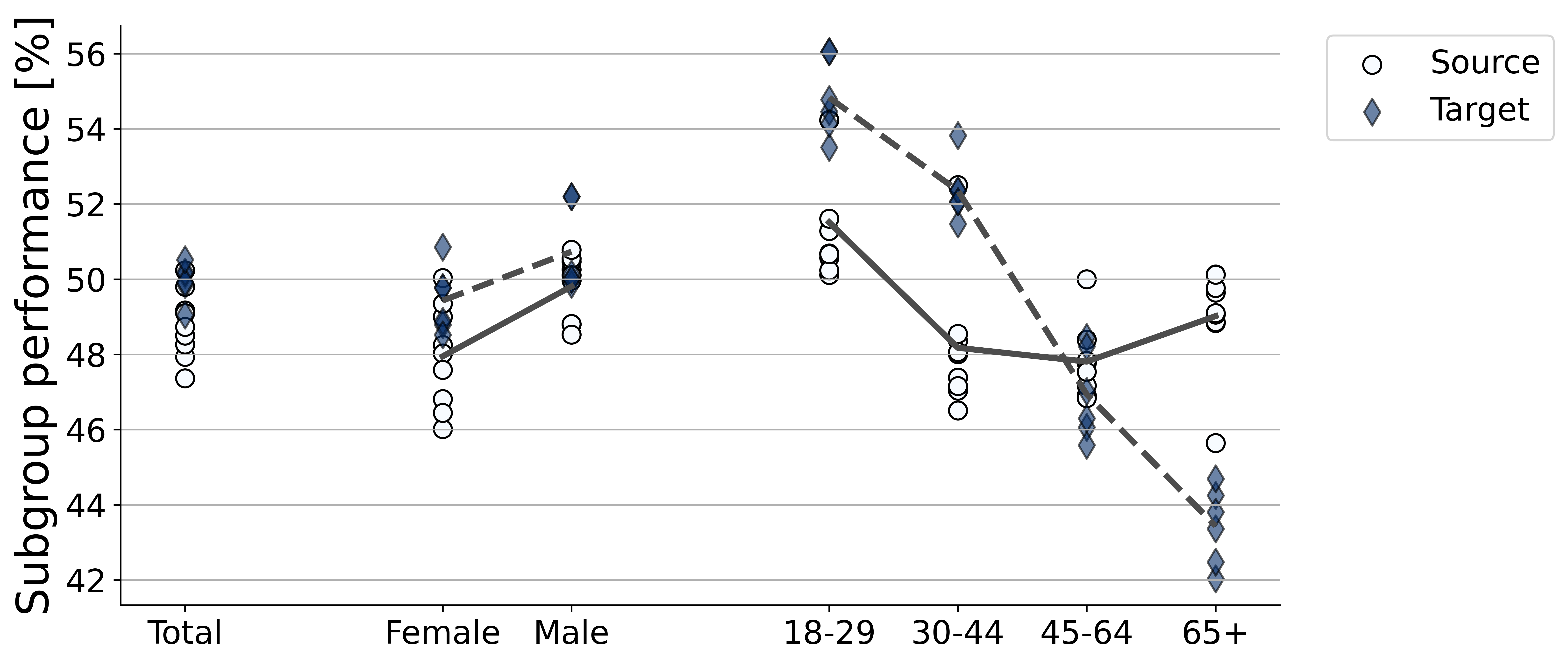}
\end{subfigure}\\
\caption{Model performance (top-1 accuracy) after joint training on the source data (circles) and target data (diamonds). (a) Original source test set. (b) Joint source test set.}
\label{fig:derm_joint_training}       
\end{figure*}

We report the results on the test set of the original source, on the test set of the joint source (randomly sampling $1,500$ cases from each dataset), and on the target set (Fig.\ref{fig:derm_joint_training}). We first observe that the overall performance gap between the source and the target has decreased. However, the fairness gap between source and target remains important, especially for age (max gap between groups: joint source = $4.33\%$, target = $11.4\%$).

We therefore conclude that joint training, as it does not introduce invariances to the environment here, does not guarantee an improvement in the transfer of fairness.

\section{Electronic health records}
\label{app:ehr}

\subsection{Dataset}
\label{app:ehr_data}

\textbf{Data availability:} The de-identified EHR data is available based on a user agreement at Physionet \citep{Goldberger2000-zg}.

\textbf{Cohort selection:} Our cohort includes patients with age $>$ 18 at time of admission and ICU length of stay (LOS) $>$ 1 day. We restrict the patient sequences to the first ICU admission, as in the MIMIC-Extract project \citep{Wang2019-ac}, and additionally restrict to only those admissions with a FIRST\_CAREUNIT defined in the MIMIC-III icustays table. This procedure results in a cohort of 28,083 patients.

\noindent \textbf{Feature and target definition:} The full feature set from \citep{Roy2021-rr} is used, leading to 31,303 binary features and 28,048 numeric features. The target is defined as a binary outcome, i.e. remaining ICU LOS $>$ 3 days, following the approach used in \citep{Wang2019-ac,Pfohl2021-xn}.

Patient demographics are used as metadata for fairness mitigation and evaluation. For illustration, we focus on sex and age (bucketed as in the dermatology application).

\noindent \textbf{Data splits:} The FIRST\_CAREUNIT defined in the MIMIC-III icustays table is used to split the dataset into a source dataset with 17,641 patients (MICU, SICU and TSICU) and target set with 10,442 patients (CCU, CSRU). The source dataset is then randomly split into training (80\%), validation (5\%), calibration (5\%) and test (10\%) sets. 

\subsection{Model and performance}
As performed in \citep{Roy2021-rr, Tomasev2019-tu, Tomasev2021-uf}, each patient's medical history is converted to a time series of one-hour aggregates including the different structured data elements (medication, labs, vitals, \dots) represented by numerical and binary variables. Based on the first 24 hourly aggregates, we predict the binary label for prolonged length of stay using a recurrent network architecture \citep{Tomasev2019-tu, Tomasev2021-uf}. We then estimate model performance in terms of accuracy across patients and per sex and age subgroups. In addition to per-group metrics, we compute demographic parity and equalized odds.

Table \ref{tab:ehr_perf} displays the detailed model performance on the source and on the target. On the source data, the model performs to $78.6 \pm 0.7 \%$ accuracy, with $\sim1\%$ performance gap between sex groups and a gap of $7.4\%$ between age subgroups. Performance increases slightly between the two environments, with the model reaching $79.7 \pm 0.9\%$. The gap between sexes increases to $2.7\%$ on the target data while it remains similar for age (7\%). In terms of fairness metrics, demographic parity is $0.002 \pm 0.002$ for sex on the source, and $0.016 \pm 0.003$ on the target. It is $0.05 \pm 0.006$ for age on the source, and $0.066 \pm 0.010$ on the target. For both attributes, we hence observe an increase in demographic parity between the source and the target. Equalized odds do not display a significant difference between the two environments, for age or sex.

\begin{table}[!h]
\centering
\caption{Model accuracy (in \%) in the source and target data, on average across model runs.}
\label{tab:ehr_perf}       
\begin{tabular}{lll}
\hline\noalign{\smallskip}
Group & Source & Target  \\
\noalign{\smallskip}\hline\noalign{\smallskip}
Total & $78.6 \pm 0.7$ & $79.7 \pm 0.9$ \\
\noalign{\smallskip}\hline\noalign{\smallskip}
Female & $77.8 \pm 0.8$ & $78.0 \pm 0.6$ \\
Male & $79.2 \pm 0.5$ & $80.7 \pm 0.5$ \\
\noalign{\smallskip}\hline\noalign{\smallskip}
$[18, 30)$ & $84.0 \pm 1.6$ & $82.5 \pm 0.9$ \\
$[30, 45)$ & $82.8 \pm 0.9$ & $84.2 \pm 0.9$ \\
$[45, 65)$ & $78.8 \pm 0.5$ & $83.2 \pm 0.6$ \\
$[65, 90)$ & $76.6 \pm 0.7$ & $77.2 \pm 0.6$ \\
\noalign{\smallskip}\hline
\end{tabular}
\end{table}

\subsection{A compound shift}
\label{app:ehr_shift}

\subsubsection{The ICU unit $S$ affects the comorbidities $M$}
We define comorbidities according to \citep{Elixhauser1998-ra, Van_Walraven2009-ei} using code in \citep{Johnson2018-zj}, obtaining a set of 30 comorbidities associated with each patient (multi-label).
In this work, we refer to comorbidities as a `summary' of ICD codes that represent the patient's medical history prior to the current admission.

\noindent \textbf{Statistical testing}: We define IPW weights based on a logistic regression that predicts the environment based on 5,000 admissions from the source and 5,000 admissions from the target data. The classifier performs with an accuracy of 58.6\% on a left-out test set comprising 20\% of the data. For each comorbidity, we assess whether its prevalence is significantly affected by the environment using weighted tests. We observe that 5 comorbidities lead to significant results (after Bonferroni correction for multiple comparisons): pulmonary circulation, peripheral vascular, liver disease, metastatic cancer and solid tumor. We however caveat this analysis by the low number of patients whom have recorded comorbidities prior to admission, leading to many tests being invalid.

\subsubsection{The ICU unit $S$ affects the treatments $T$}

\noindent \textbf{Definition of treatments}:
We identify treatments based on the `MedicationRequest' field in the FHIR representation \citep{Rajkomar2018-oy} of MIMIC. For each of those treatments, we construct a summary by counting the number of hourly requests in the first 24 hours after admission. This summary is used for illustrating specific medications (beta-blockers, vasopressors/inotropes and ACE inhibitors) and to control for treatments in section \ref{app:ehr_label_shift}. Given that this representation includes 5,873 dimensions, we build a further 1-dimensional summary \citep{Lipton2018-ho} by training a model that predicts LOS from the time series of the medication requests on the source training split. This model reaches an accuracy of 76.9\% (AUPRC: 0.54, AUROC: 0.75) on the source test split.

\noindent \textbf{Statistical testing}: We assess whether treatments are directly affected by $S$ by computing weights to balance $\{A, M\}$. We note that this analysis is limited as comorbidities are not reported for all patients. Nevertheless, a binary model discriminating between environments based on $\{A,M\}$ leads to 62\% accuracy. Given the dimensionality of treatments, we perform a weighted test on the summary of treatments. Considering all medication, we observe that patients receive different treatments in different units (weighted test, $p=0.015$), hence $T \nind S | \{A,M\}$.

\begin{figure*}[!ht]
\centering
\begin{subfigure}[t]{0.03\textwidth}
\textbf{a}
\end{subfigure}
\begin{subfigure}[t]{0.6\textwidth}
\includegraphics[width=\linewidth,valign=t]{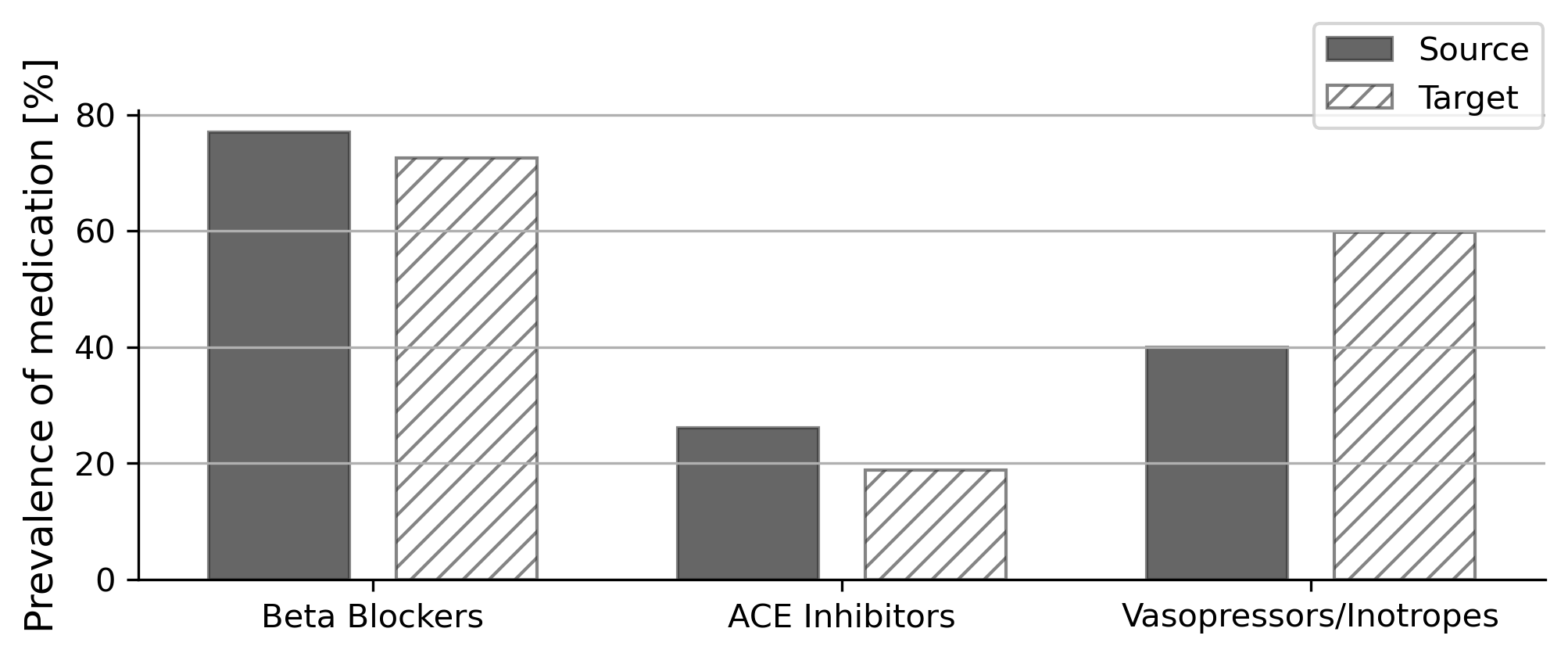}
\end{subfigure}\\
\begin{subfigure}[t]{0.03\textwidth}
\textbf{b}
\end{subfigure}
\begin{subfigure}[t]{0.6\textwidth}
\includegraphics[width=\linewidth,valign=t]{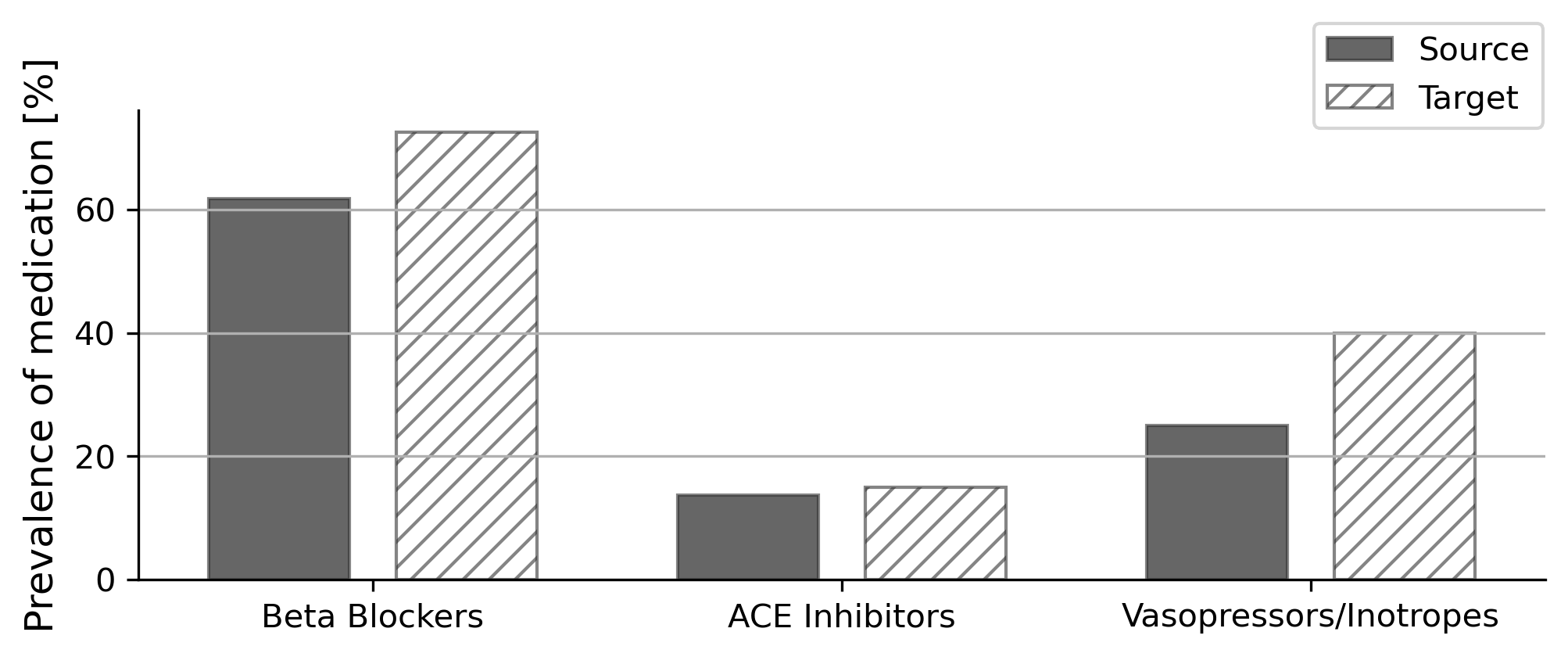}
\end{subfigure}\\
\caption{Per comorbity  prevalence of treatment (in \%). (a) Peripheral vascular comorbidities. (b) Solid tumor.}
\label{fig:ehr_data_com}       
\end{figure*}

As an illustration, we estimate what proportion of patients receives treatments such as beta-blockers, vasopressors/inotropes and/or ACE inhibitors (Figure~\ref{fig:ehr_data_com}) for patients with (a) vascular comorbidities, and (b) tumor history. 

\subsubsection{The ICU unit $S$ does not affect the length of stay $Y$}
\label{app:ehr_label_shift}

We refer to the summary of treatments of dimension 5,873 built above to define $T$ and build the feature set $\{A, M, T\}$ to estimate overlap weights such that we can test whether $P(Y | A, M, T, S=0) = P(Y | A, M, T, S=1)$. The treatment features with non-null values are normalized to be in the range [0, 1] before we build a classifier that distinguishes between the two environments. The model achieves $87.2\%$ accuracy in predicting the environment. We derive the weights for each data point in the two environments and assess with a weighted t-test whether the prevalence in the two environments is similar. Our results show that the ICU unit might not affect the prevalence of the length of stay ($p=0.18$). We however note that our analysis is limited by the large dimensionality of $\{A, M, T\}$ to build the weights. Indeed, the gradient boosted tree model $P(S | A, M, T)$ can then rely on many different signals to build its predictions \citep{DAmour2020-bz} and display high variance across bootstraps. This would result in an under-powered test.



\subsubsection{The ICU unit $S$ does not affect the labs and vitals $X$}
Based on the causal graph, we use the same IPW weights as in Section~\ref{app:ehr_label_shift}. We build a summary of $X$ by predicting $Y$ from all features that are not comorbidities, demographics or treatments. This recurrent model reaches 77.84\% accuracy (AUPRC: 0.62, AUROC: 0.80) on the source test split. Based on this summary, we observe that labs and vitals do not display evidence of being affected by the environment ($p=0.45$).



\end{document}